\documentclass{article}

\usepackage{arxiv}

\usepackage[utf8]{inputenc} 
\usepackage[T1]{fontenc}    
\usepackage{hyperref}       
\usepackage{url}            
\usepackage{booktabs}       
\usepackage{setspace}
\usepackage{amsfonts}       
\usepackage{nicefrac}       
\usepackage{microtype}      
\usepackage{graphicx}
\usepackage{natbib}
\usepackage{doi}
\usepackage{enumerate}
\usepackage{float}
\usepackage{algorithm}      
\usepackage{algpseudocode}  
\usepackage{algcompatible} 
\usepackage{algpseudocode}
\usepackage{enumitem}
\usepackage{xcolor}         
\usepackage{graphicx}
\usepackage{amsthm}
\usepackage{amsmath}
\usepackage{caption}
\usepackage[capitalise]{cleveref}
\usepackage{amssymb}
\usepackage{bbm}
\usepackage{bm}
\usepackage{acronym}
\usepackage{enumitem}
\usepackage{tikz-cd}
\usepackage{tikz}
\usetikzlibrary{arrows,positioning,patterns,shapes,decorations}

\usetikzlibrary{matrix}
\usepackage{booktabs}
\newtheorem{theorem*}{Theorem}
\newtheorem{definition*}{Definition}

\newtheorem{lemma*}{Lemma}

\DeclareMathOperator*{\argmin}{arg\,min}

\definecolor{lowcolor}{rgb}{0.43, 0.21, 0.1}
\definecolor{highcolor}{rgb}{0.0, 0.5, 0.0}

\title{Causal Optimal Transport of Abstractions}

\author{
  Yorgos Felekis \\
  University of Warwick \\
  \texttt{yorgos.felekis@warwick.ac.uk} \\
   \And
  Fabio Massimo Zennaro \\
  University of Bergen \\
  \texttt{fabio.zennaro@uib.no} \\
   \And
   Nicola Branchini \\
  University of Edinburgh \\
  \texttt{n.branchini@sms.ed.ac.uk} \\
  \And
   Theodoros Damoulas \\
  University of Warwick \\
  \texttt{t.damoulas@warwick.ac.uk} 
  }

\DeclareMathSymbol{\mh}{\mathord}{operators}{`\-}

\newcommand{\dom}[1]{\operatorname{dom}[#1]}
\newcommand{\scmdo}{\operatorname{do}}

\newcommand{\scmbase}{\mathcal{M}}
\newcommand{\scmabst}{\mathcal{M}^\prime}
\newcommand{\intervsetbase}{\mathcal{I}}
\newcommand{\intervsetabst}{\mathcal{I}^\prime}
\newcommand{\structfuncset}{\mathcal{F}}
\newcommand{\scmdag}{\mathcal{G}}

\newcommand{\tauomega}{\tau\mh\omega}
\newcommand{\intervpairs}{\Pi_{\omega}}

\newcommand{\cmpt}[2]{\operatorname{Cmp}(#1,#2)}

\newcommand{\prob}{\mathbb{P}}
\newcommand{\expval}{\mathbb{E}}

\newcommand{\onevector}[1]{\mathbbm{1}_{#1}}
\newcommand{\indicator}{\mathbbm{1}}
\newcommand{\logicaland}{\wedge}
\newcommand{\cardinality}[1]{\left|#1\right|}

\newcommand{\transpose}{\top}
\newcommand{\cota}{{\texttt{COTA}}}
\newcommand{\scm}[1]{{\texttt{SCM}#1}}
\newcommand{\ot}{{\texttt{OT}}}
\newcommand{\dirag}[1]{{\texttt{DAG}#1}}
\newcommand{\baryOT}{{\texttt{Bary OT}}}

\newcommand{\pairOT}{{\texttt{Pwise OT}}}
\newcommand{\mapOT}{{\texttt{Map OT}}}

\newcommand{\fro}{{\texttt{FRO}}}
\newcommand{\jsd}{{\texttt{JSD}}}
\newcommand{\wass}{{\texttt{WASS}}}

\newcommand{\ca}{{\texttt{CA}}}

\begin{document}

\maketitle
\begin{abstract}
    Causal abstraction (\ca{}) theory establishes formal criteria for relating multiple structural causal models (\scm{s}) at different levels of granularity by defining maps between them. These maps have significant relevance for real-world challenges such as synthesizing causal evidence from multiple experimental environments, learning causally consistent representations at different resolutions, and linking interventions across multiple \scm{s}. In this work, we propose \cota{}, the first method to learn abstraction maps from observational and interventional data without assuming complete knowledge of the underlying \scm{s}. In particular, we introduce a multi-marginal Optimal Transport (\ot{}) formulation that enforces \emph{do-calculus} causal constraints, together with a cost function that relies on interventional information. We extensively evaluate \cota{} on synthetic and real world problems, and showcase its advantages over non-causal, independent and aggregated \ot{} formulations. Finally, we demonstrate the efficiency of our method as a data augmentation tool by comparing it against the state-of-the-art \ca{} learning framework, which assumes fully specified \scm{s}, on a real-world downstream task.
\end{abstract}


\section{Introduction}\label{sec:Introduction} 

Learning relations between models and underlying representations at different levels of granularity is a key challenge across sub-fields of AI as it can enable aggregation of information, transfer learning, emulation via surrogate models, and multi-scale estimation and reasoning e.g. \citep{weinan2011principles,somnath2021multi,geiger2021causal}. Rigorous relationships of abstraction between such models would enable utilising seemingly incompatible data, leading to improved inferences via evidence synthesis and cost savings by minimising the need for extensive data collection.

In causality the notion of abstraction is fundamental for causal representation learning, where causal variables might be abstractions of underlying quantities or when relations are sought between micro- and macro-level models of the same underlying process \citep{scholkopf2021toward,chalupka2017causal}. Relations between causal models and estimands across multiple environments have been studied under transportability \citep{pearl2011transportability} and multi-environment causal analysis \citep{peters2016causal,yin2021optimization}. A theory of causal abstraction has been formalised \citep{rubenstein2017causal,beckers2018abstracting,rischel2020category,massidda2022causal} through the definition of a map relating two causal models representing the same system in different levels of detail and a measure of interventional consistency evaluating the discrepancy between the two under interventions \citep{beckers2020approximate,rischel2020category}. This framework has been used in the field of explainability \citep{geiger2021causal}, where, given an abstraction, a neural network is trained to behave consistently with an abstracted model. While limited work exists on abstraction learning, \citep{zennaro2023jointly} proposed a differentiable programming solution to learn an abstraction between two causal models in the $\mathbf{\alpha}$-abstraction framework of \cite{rischel2020category}, but with the strong assumption that the underlying causal models were fully specified. In this work, we lift this restrictive assumption of complete \scm{} knowledge and study the more realistic setting in which the information available to the modeler is the graph underlying the causal model together with observational and interventional data. We make the following contributions:
\begin{itemize}[itemsep=0.01pt]
    \item We formalise the problem of learning causal abstractions (\ca{}) from observational and interventional data in the $(\tau, \omega)$-framework \citep{rubenstein2017causal}. 
    \item We introduce the first method to address this problem without assuming full knowledge of the underlying causal models and showcase its superiority against our multiple developed baselines and  prior work \citep{zennaro2023jointly} of alternative \ca{} learning frameworks that also assumes fully specified \scm{s}.
    \item To do so, we develop a causal Optimal Transport (\ot) formulation for abstraction learning, named \cota{}, where observational and interventional distributions of the base and abstracted models act as marginals in a \cite{kantorovich1942} joint \ot{} problem with multiple transport plans. Further, we prove the joint convexity of \cota{} in the induced plans, guaranteeing an optimal solution to the optimization problem.
    \item We incorporate causal knowledge to the optimisation problem by introducing \emph{do-calculus} constraints and a causally informed cost function. We demonstrate that this enables us to learn better abstraction maps compared to non-causal or independent solutions and also makes \cota{} a potent data augmentation tool.
\end{itemize}

\section{Background on causality, abstractions, and optimal transport}\label{sec:Background} 
In this section we introduce basic definitions from the field of causality, causal abstractions, and optimal transport. We use the following standard notation to formalise causal models: boldface capital $\mathbf{X}$ denotes a set of random variables, and capital letter $X_i$ denotes the $i$-th random variable in $\mathbf{X}$; boldface small $\mathbf{x}$ denotes a set of values realising $\mathbf{X}$; $x_i$ denotes the $i$-th value in $\mathbf{x}$. We use boldface $\mathbb{P}$ to refer to the underlying probability measures.

\subsection{Causality}\label{sec:causality}
\begin{definition*}[\scm{} \citep{pearl2009causality}]\label{def:scm}
    A structural causal model $\scmbase$ is a tuple $\langle \mathbf{X},\mathbf{U},\structfuncset,\prob(\mathbf{U}) \rangle$, where 
$\mathbf{X}$ is a set of $N$ endogenous random variables, each one with domain $\dom{X_i}$, $1\leq i \leq N$; $\mathbf{U}$ is a set of exogenous random variables each associated with an endogenous variable; $\structfuncset$ is a set of structural functions, one for each endogenous variable $X_i\in\mathbf{X}$ defined as $f_i:\dom{\operatorname{PA}(X_i)}\times \dom{U_i}\to \dom{X_i}$ where $\operatorname{PA}(X_i)\subseteq \mathbf{X} \setminus X_i$; 
 $\prob(\mathbf{U}) = \prod_{i=1}^N \prob(U_i)$ is a joint probability distribution over $\mathbf{U}$. 
\end{definition*}
We make a few standard assumptions on our \scm{s}. We assume \emph{acyclicity}, implying that the \scm{} $\scmbase$ entails a directed acyclic graph (\dirag{}) $\scmdag_\scmbase$ where nodes correspond to the endogenous variables $\mathbf{X}$ and edges are defined by the signature of the functions in $\structfuncset$ \citep{peters2017elements}. \footnote{Also, assuming the \emph{measurability} of the structural functions in $\structfuncset$ we can derive, via a pushforward over the functions in $\structfuncset$, the probability distribution $\prob_{\scmbase}(\mathbf{X})$ over the endogenous variables.}
We will also assume \emph{faithfulness}, guaranteeing that independencies in the data are captured in the graphical model, and \emph{causal sufficiency}, meaning that there are no unobserved confounders \citep{spirtes2000causation}.
\begin{definition*}[Interventions \citep{pearl2009causality}]
Given a \scm{} $\scmbase$, an (exact) intervention $\iota = \scmdo(\mathbf{A}=\mathbf{a})$, where for each endogenous variable $A_i \in \mathbf{A}\subseteq\mathbf{X}$ we have a value $a_i \in \mathbf{a}$ and $a_i \in \dom{A_i}$, is an operator that replaces each function $f_i$ associated with the variable $A_i$ with the constant $a_i$.
\end{definition*}
Graphically, the intervention $\iota = \scmdo(\mathbf{A}=\mathbf{a})$ mutilates the induced graph $\scmdag_\scmbase$ by removing the incoming arrows in each node $A_i$ and replacing $f_i$ with the constant $a_i$. In this way, an intervention defines a new \emph{post-intervention} \scm{} $\scmbase_{\iota}$ described by the probability distribution $\prob_{\scmbase_{\iota}}(\mathbf{X})$. Whenever clear from the context, we shorthand $\scmdo(\mathbf{A}=\mathbf{a})$ to $\scmdo(\mathbf{a})$.  

Also, sets of interventions are equipped with a natural partially-ordered set (poset) structure\footnote{A partially-ordered set (poset) is a pair $(\mathcal{S}, \preceq)$, with a non-empty set $\mathcal{S}$ and reflexive, anti-symmetric, and transitive  relation $\preceq$. In a poset, elements are comparable if one precedes the other. Totally-ordered sets are posets, where all their element pairs are comparable.} with respect to containment: given $\iota = \scmdo(\mathbf{a})$ and $\eta = \scmdo(\mathbf{b})$, $\iota \preceq \eta$ iff $\mathbf{A} \subseteq \mathbf{B}$ and whenever $B_j = A_i$ then $b_j = a_i$ \citep{rubenstein2017causal}. 
\begin{definition*}[Compatibility]
    Given a set of values $\mathbf{b} \in \dom{\mathbf{B}}$, $\mathbf{B} \subseteq \mathbf{X}$, and an intervention $\iota = \scmdo(\mathbf{a})$ we say that $\mathbf{b}$ and $\iota$ are \emph{compatible} $\cmpt{\mathbf{b}}{\iota}$ if $\scmdo(\mathbf{a}) \preceq \scmdo(\mathbf{b})$.
\end{definition*}
Thus, a set of values $\mathbf{b}$ such that $\cmpt{\mathbf{b}}{\iota}$ is a set of values that agrees with the intervention $\iota$; a set of values $\mathbf{b}$ for which it does not hold $\cmpt{\mathbf{b}}{\iota}$, is a setting of $\scmbase$ that is ruled out by $\iota$.

\subsection{Causal Abstractions} 
Causal abstractions formalise relations between low-level (base) and high-level (abstracted) models, enabling causal evidence synthesis and consistent representation learning among them. This allows shifting between varying levels of granularity based on the specific inquiry or available data.
\begin{definition*}[$\tauomega$ Exact Transformation \citep{rubenstein2017causal}]\label{def:tauomega}
Given a base model $\scmbase$ and an abstracted model $\scmabst$ respectively equipped with posets $\intervsetbase$, $\intervsetabst$ of interventions, and a surjective and order-preserving map $\omega: \intervsetbase \to \intervsetabst$, a $\tauomega$ transformation is a map $\tau: \dom{\mathbf{X}} \to \dom{\mathbf{X^{\prime}}}$. An exact transformation is a map $\tau$ such that 
\begin{equation}\label{eq:tau_omega_def}
\tau_{\#} \left(\prob_{\scmbase_{\iota}}(\mathbf{X})\right) = \prob_{\scmabst_{\omega(\iota)}}(\mathbf{X^{\prime}}), ~\forall \iota \in \intervsetbase.
\end{equation}
\end{definition*}
For the $\omega$ map, \emph{order-preserving} implies that $\iota \preceq \eta \implies \omega(\iota)\preceq \omega(\eta)$ and \emph{surjective} that $\forall \iota^{\prime} \in \intervsetabst ~ \exists ~ \iota \in \intervsetbase$ such that $\omega(\iota) = \iota^{\prime}$. An exact $\tauomega$ transformation is a form of abstraction between probabilistic causal models \citep{beckers2020approximate} that ensures commutativity between interventions and transformations: intervening via $\iota$ and then abstracting via $\tau$ leads to the same result as abstracting first via $\tau$ and then intervening via $\omega(\iota)$. Exactness is rare in realistic scenarios due to approximation and uncertainty. Thus, we permit approximate transformations \citep{beckers2020approximate,rischel2021compositional} and introduce the concept of average abstraction error.
\begin{definition*}[Abstraction error]\label{def:abstractionerr}
Let $\tau$ be a $\tauomega$ transformation between \scm{} $\scmbase$ and $\scmabst$ wrt $\intervsetbase$ and $\omega$. Given a discrepancy measure $\mathcal{D}$ between distributions, and a distribution $q$ over the intervention set $\intervsetbase$, we evaluate the approximation introduced by $\tau$ as the \emph{abstraction error}:
\begin{equation}\label{eq:abst_error}
e(\tau) = \expval_{\iota \sim q }\left[ \hspace{0.7mm} \mathcal{D}\left( \tau_{\#} (\prob_{\scmbase_{\iota}}), ~ \prob_{\scmabst_{\omega(\iota)}}\right) \hspace{0.7mm} \right]
\end{equation}
\end{definition*} 
We assume a uniform distribution $q$ over $\intervsetbase$, treating each intervention as equally important. However, a modeller may modify this distribution, assigning varying importance to interventions of particular interest. \cref{fig:tau_omega} (left) shows the commutative diagram induced by such an approximate abstraction.

\subsection{Optimal Transport}\label{sec:optimal_transport}
Optimal Transport theory \citep{villani2009optimal} provides a mathematical framework to efficiently redistribute probability mass between distributions by minimising a cost function. Consider two probability measures $\prob_{\scmbase}(\mathbf{X}), \prob_{\scmabst}(\mathbf{X}^{\prime})$ on domains $\dom{\mathbf{X}},\dom{\mathbf{X}^{\prime}}$. When only samples from the measures are available, computational \ot{} resorts to the corresponding empirical measures of them \citep{peyre2019computational}, say $\widehat{\prob}_{\scmbase}(\mathbf{X}) = \alpha, \widehat{\prob}_{\scmabst}(\mathbf{X}^{\prime})=\beta$. Thus, we obtain i.i.d. data from the distributions, $\{ \mathbf{x}_{j} \}_{j=1}^{N} \sim \prob_{\scmbase}(\mathbf{X})$, $\{ \mathbf{x}_{i}^{\prime} \}_{i=1}^{M} \sim \prob_{\scmabst}(\mathbf{X}^{\prime})$ and construct the empirical measures as $\alpha = \sum_{j=1}^{N} \alpha_j \delta_{\mathbf{x}_{j}}$, $ \beta = \sum_{i=1}^{M} \beta_i \delta_{\mathbf{x}_{i}^{\prime}}$ where $\delta$ is the Dirac measure. Note that in principle $\dom{\mathbf{X}},\dom{\mathbf{X}^{\prime}}$ could be either continuous or discrete. 

The \cite{monge1781memoire} formulation of \ot{} then aims to find a map  $T: \dom{\mathbf{X}} \to \dom{\mathbf{X}^{\prime}}$ that pushforwards $\alpha$ onto $\beta$, the one that minimises: 
\begin{equation}\label{eq:monge}
T^{\star} = \argmin_{T:~ T_{\#}\alpha = \beta}\sum_{j=1}^Nc(x_j,T(x_j))
\end{equation}
where $c:\mathbf{X} \times \mathbf{X^{\prime}} \to \mathbb{R}_{\geq 0}$ represents the cost of moving a unit mass from $\mathbf{X}$ to $\mathbf{X}^{\prime}$. Due to the challenge of unique solution existence of \cref{eq:monge}, \cite{kantorovich1942} introduced a more flexible formulation that seeks to find a coupling matrix, defined as an element of the set of stochastic matrices with given marginals, i.e., $
\mathcal{U}(\alpha,\beta)= \{P\in \mathbb{R}_{\geq 0}^{M \times N}: P\onevector{M}=\alpha,P^\transpose \onevector{N}=\beta\} $.
The set $\mathcal{U}(\alpha,\beta)$ is bounded and defined by $M+N$ equality constraints, and
therefore is a convex polytope \citep{peyre2019computational}. Solving the induced \ot{} problem directly often poses significant computational challenges. To address this issue, \emph{Entropic} \ot{}, an efficient and tractable formulation of \ot{} which incorporates an entropic regularization term, is usually used in practice . This is formalised as:
\begin{align}\label{eq:entropic_ot}
P^{\star} = \ot_c(\alpha,\beta)=\argmin_{P\in\mathcal{U(\alpha,\beta)}}\left<C,P\right>-\epsilon\mathcal{H}(P) =\argmin_{P\in\mathcal{U(\alpha,\beta)}}\sum_{i=1, j=1}^{M, N} C_{i, j} P_{i, j} -\epsilon\mathcal{H}(P) 
\end{align}
where $\left<\cdot,\cdot\right>$ denotes the Frobenius inner product, $C \in  \mathbb{R}_{\geq 0}^{M \times N}$ is a cost matrix where each element is constructed with the \ot{} cost function $C_{i,j} = c(\mathbf{x}_{i}^{\prime}, \mathbf{x}_{j})$ and $\mathcal{H}(P)$ is the discrete entropy of the coupling matrix $P$ with $\epsilon>0$ a trade-off parameter.
The original Kantorovich \ot{} problem is now a special case of \cref{eq:entropic_ot} when the entropic regularization parameter $\epsilon$ is set to 0. Further details on Optimal Transport, can be found in \cref{sec:ot_appendix_content}.
 
\section{Abstraction Learning as Multi-marginal Optimal Transport} \label{sec:probstatement}
\begin{figure}
        \centering
        \begin{tikzpicture}[shorten >=1pt, auto, node distance=1cm, thick, scale=0.8, every node/.style={scale=0.8}]
        \tikzstyle{node_style} = []
        
        \node[node_style,lowcolor] (P_MA) at (-6.5,3) {$\prob_{\scmbase}$};
        \node[node_style,lowcolor] (P_MAi) at (-2.2,3) {$\prob_{{\scmbase}_{\iota}}$};
        \node[node_style,highcolor] (P_MB) at (-6.5,0) {$\prob_{\scmabst}$};
        \node[node_style,highcolor] (P_MBi) at (-3.5,0) {$\prob_{{\scmabst}_{\omega(\iota)}}$};
        \node[node_style,highcolor] (P_MBii) at (-2.2,1) {$\tau_\#(\prob_{{\scmbase}_{\iota}})$};
        
        \draw[->]  (P_MA) to node[above,font=\small]{$\iota$} (P_MAi);
        \draw[->]  (P_MA) to node[left,font=\small]{$\tau$} (P_MB);
        \draw[->]  (P_MB) to node[below,font=\small]{$\omega(\iota)$} (P_MBi);
        \draw[->]  (P_MAi) to node[right,font=\small]{$\tau$} (P_MBii); 
        \draw[dotted,bend right]  (P_MBi) to node[right,font=\small]{$\mathcal{D}(\cdot, \cdot)$} (P_MBii);
        
        \node[node_style,lowcolor] (P_M0) at (0.5,0) {$\emptyset$};
        \node[node_style,lowcolor] (P_M0i1) at (2,1) {$\iota_1$};
        \node[node_style,lowcolor] (P_M0i2) at (-1,2) {$\iota_2$};
        \node[node_style,lowcolor] (P_M0i3) at (0.5,3) {$\iota_3$};
        
        \node[node_style,highcolor] (P_M1) at (4.5,0) {$\emptyset$};
        \node[node_style,highcolor] (P_M1k1) at (4.5,1.5) {$\eta_1$};
        \node[node_style,highcolor] (P_M1k2) at (4.5,3) {$\eta_2$};
        
        \draw[->]  (P_M0) to (P_M0i1);
        \draw[->]  (P_M0) to (P_M0i2);
        \draw[->]  (P_M0i1) to (P_M0i3);
        \draw[->]  (P_M0i2) to (P_M0i3);
        
        \draw[->]  (P_M1) to (P_M1k1);
        \draw[->]  (P_M1k1) to (P_M1k2);
        
        \draw[->,dotted,red]  (P_M0) to node[below,font=\small]{$\omega(\emptyset) = \emptyset$} (P_M1);
        \draw[->,dotted,red]  (P_M0i1) to node[below,font=\small]{$\omega(\iota_1) = \eta_1$} (P_M1k1);
        \draw[->,dotted,red]  (P_M0i2) to node[above,font=\small]{$\omega(\iota_2) = \eta_1$} (P_M1k1);
        \draw[->,dotted,red]  (P_M0i3) to node[above,font=\small]{$\omega(\iota_3) = \eta_2$} (P_M1k2);
        
        \node[node_style,lowcolor] (P_M0) at (7.5,0) {$\prob_{\scmbase}$};
        \node[node_style,lowcolor] (P_M0i1) at (9,1) {$\prob_{\scmbase_{\iota_1}}$};
        \node[node_style,lowcolor] (P_M0i2) at (6,2) {$\prob_{\scmbase_{\iota_2}}$};
        \node[node_style,lowcolor] (P_M0i3) at (7.5,3) {$\prob_{\scmbase_{\iota_3}}$};
        
        \node[node_style,highcolor] (P_M1) at (11,0) {$\prob_{\scmabst}$};
        \node[node_style,highcolor] (P_M1k1) at (11,1.5) {$\prob_{\scmabst_{\eta_1}}$};
        \node[node_style,highcolor] (P_M1k2) at (11,3) {$\prob_{\scmabst_{\eta_2}}$};
        
        \draw[->]  (P_M0) to node[below,font=\small]{$\iota_1$} (P_M0i1);
        \draw[->]  (P_M0) to node[left,font=\small]{$\iota_2$} (P_M0i2);
        \draw[->]  (P_M0i1) to node[left,font=\small]{$\iota_3$} (P_M0i3);
        \draw[->]  (P_M0i2) to node[below,font=\small]{$\iota_4$} (P_M0i3);
        
        \draw[->]  (P_M1) to node[left,font=\small]{$\eta_1$} (P_M1k1);
        \draw[->]  (P_M1k1) to node[left,font=\small]{$\eta_2$} (P_M1k2);
        
        \draw[->,dotted,red]  (P_M0) to node[below,font=\small]{$\tau$} (P_M1);
        \draw[->,dotted,red]  (P_M0i1) to node[below,font=\small]{$\tau$} (P_M1k1);
        \draw[->,dotted,red]  (P_M0i2) to node[above,font=\small]{$\tau$} (P_M1k1);
        \draw[->,dotted,red]  (P_M0i3) to node[above,font=\small]{$\tau$} (P_M1k2);
        \end{tikzpicture}
    \captionsetup{skip=10pt}
    \caption{{\textbf{(Left)} Abstraction commutative diagram. We run two different paths: \textbf{(a)} apply $\iota$ on the base model $\scmbase$, and then $\tau$; or \textbf{(b)} apply $\tau$ to get the abstracted model $\scmabst$, and then $\omega(\iota)$. We compute the distance between $\tau_{\#} (\prob_{\scmbase_{\iota}})$ and $\prob_{\scmabst_{\omega(\iota)}}$ using a divergence $\mathcal{D}$, indicated by the dotted line. If $\mathcal{D}=0$, the commutative diagram expresses an exact $\tauomega$ abstraction. \textbf{(Middle)} The action of the $\omega$ map on a poset of interventions $\intervsetbase = \{ \emptyset, \iota_1, \iota_2, \iota_3,\}$. The map $\omega$ is a surjective order-preserving map from $\intervsetbase$ to $\intervsetabst$.
    \textbf{(Right)} The action of $\tau$ map on a poset of distributions induced by $\intervsetbase$. A single map $\tau$ pushforwards all the base distributions $\prob_{\scmbase_\iota}$ onto abstracted distributions $\prob_{\scmabst_\eta}$.}}
    \label{fig:tau_omega}
\end{figure}
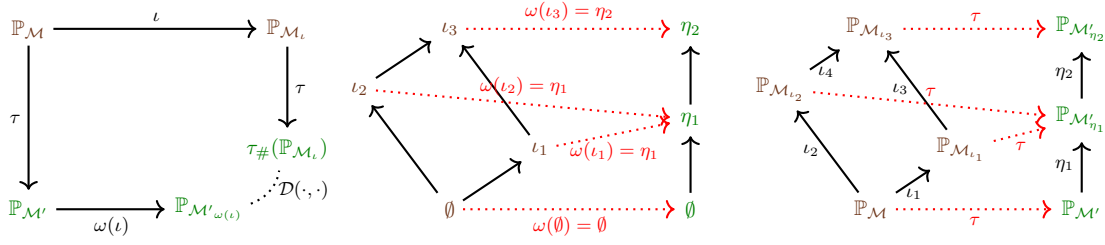

In this section we formalise the \ca{} learning problem from data as a multi-marginal \ot{} problem, and show how we inject causal information into it. We assume \textbf{(a)} access to the causal \dirag{}s of the base $\scmbase$ and the abstracted $\scmabst$ models; \textbf{(b)} a finite set of interventions $\intervsetbase$; \textbf{(c)} an intervention mapping $\omega: \intervsetbase \to \intervsetabst$; \textbf{(d)} samples from the observational and interventional distributions. 
Thus, each base model intervention $\iota \in \intervsetbase$ and its image $\omega(\iota)\in \intervsetabst$ yield a pair of empirical distributions, denoted as $\pi_{\iota} = \{ (\widehat{\prob}_{\scmbase_{\iota}}(\mathbf{X}), \widehat{\prob}_{\scmabst_{\omega(\iota)}}(\mathbf{X}^{\prime})) \}$. We define the set of all these pairs as $\intervpairs(\intervsetbase)$; see \cref{fig:tau_omega}(middle) for the structure of $\intervpairs(\intervsetbase)$.
Our aim is to learn a single $\tauomega$  transformation from data sampled from the pairs in $\intervpairs(\intervsetbase)$. We address this challenge by viewing each pair $\pi_{\iota}$ as marginals in an Entropic \ot{} problem within the Kantorovich formulation for discrete measures\footnote{The Kantorovich framework is essential for abstraction problems where marginal distribution dimensions mismatch, necessitating mass splitting between base and abstract points, which renders Monge maps infeasible.}. We compute a plan $P^{\iota}$ for each pair $\pi_{\iota}$, thereby leading to a \emph{multi-marginal} optimization problem, made up of $|\intervpairs(\intervsetbase)|$ independent \ot{} problems: 
\begin{align}\label{eq:optproblem}
    \bm{P}^{\star} = \ot_c(\intervpairs(\intervsetbase)) = \argmin_{\substack { \{ P^{\iota} \in \mathcal{U}(\pi_{\iota})} \}_{\iota \in \intervsetbase} }
\left\{\sum_{\iota \in \intervsetbase}\Big<C,P^{\iota}\Big> -\epsilon\mathcal{H}(P^{\iota})\right\}
\end{align}
where $\mathcal{U}(\pi_{\iota})$ is the transport polytope of each pair $\pi_{\iota}$. As shown in \cref{fig:tau_omega} (right), since we are looking for a single transformation $\tau$, the plans $P^{\iota}$ obtained by solving \cref{eq:optproblem} have to be aggregated into a single plan $\bar{P}$, from which the map $\tau$ can be derived. In our context, we compute the final $\tau$ as a stochastic mapping $f_s: \dom{\mathbf{X}} \to \mathcal{A}^{|\dom{\mathbf{X^{\prime}}}|}$, induced from $P$, where $\mathcal{A}^n = \left\{\bm{p} \in \mathbb{R}^n, :~ p_i \geq 0, \sum_i p_i =1 \right\}$ the simplex in $\mathbb{R}^{n}$, to account for uncertainty of the learned abstraction with $n<\infty$ in order to allow the computation of the probability vectors. The stochastic mapping converts the mass allocation, induced by $\bar{P}$, by assigning each base sample to a probability vector, depicting a distribution over the abstracted samples.
\paragraph{Introducing causal knowledge into \ot{}.}
The optimization problem of \cref{eq:optproblem} is a collection of independent \ot{} problems. However, we know that the marginals of different plans that correspond to the same model are linked since they are interventional distributions of the same \scm{} and could be formally related via \emph{do-calculus} operations. Furthermore, standard costs used in \ot{} (e.g. $l_1,~ l_2,~l_p$) cannot capture a meaningful notion of distance between the domain of the base and the abstracted model. For these reasons, we enrich \cref{eq:optproblem} in two ways: \textbf{(a)} by establishing structural causal constraints amongst the different plans able to capture the relation of their marginals and \textbf{(b)} by introducing causal knowledge through the definition of a suitable cost function which integrates knowledge from the $\omega$ map. We show how such an enrichment transforms the initial problem into a joint causally informed multi-marginal optimization problem.

\section{Methodology}
In this section we present our methodology: \cref{sec:constraints} shows how \emph{do-calculus} \citep{pearl2009causality} constraints can be incorporated into the optimization problem; \cref{sec:omegacost} defines a meaningful cost for the \ot{} problem;  \cref{sec:cota} presents the end-to-end \hyperref[sec:cota_alg]{\cota{} algorithm} and analyzes its convexity and computational complexity. 

\subsection{\emph{Do-calculus} constraints for optimal transport of abstractions}\label{sec:constraints}
As mentioned in \cref{sec:probstatement}, marginals of different plans can be related via \emph{do-calculus} when they refer to the same \scm{}. We first analyze the intervention set structure to identify comparable interventions.
\begin{definition*}[(Maximal) Chain] Given a poset $\intervsetbase$, a \emph{chain} $\intervsetbase_q$ is a totally ordered subset of $\intervsetbase$. Let $\mathcal{C}(\intervsetbase) = \{\intervsetbase_1,\ldots,\intervsetbase_Q\}$ be the set of all chains for $\intervsetbase$.
A chain $\intervsetbase_q \in \mathcal{C}(\intervsetbase)$ is \emph{maximal} if $\neg\exists$ chain $\intervsetbase_s\in\mathcal{C}(\intervsetbase)$ such that $\intervsetbase_q \subset \intervsetbase_s$.
\end{definition*}
Interventions $\iota,\eta \in \intervsetbase$ are comparable $\iota \preceq_{\intervsetbase_q} \eta$ if there exists at least one chain $\intervsetbase_q$ to which they both belong. Notice that comparability in $\intervsetbase$ extends immediately to $\intervsetabst$ because of the order-preservation of $\omega$. Further, we also extend the notion of comparability to transport plans.
\begin{definition*}[Comparable plans]
    Let intervention pairs of distribution $\pi_{\iota}, \pi_{\eta}$ for $\iota, \eta \in \intervsetbase$. The induced transport plans $P^{\iota}, P^{\eta}$ are \emph{comparable} $P^{\iota} \preceq P^{\eta} \text{ iff } \exists ~ \intervsetbase_q \in \mathcal{C}(\intervsetbase) \text{ such that } \iota \preceq_{\intervsetbase_q} \eta$.
\end{definition*}
Marginals of comparable plans can be related via \emph{do-calculus}. Let $\iota  = \scmdo(\mathbf{a}), \omega(\iota) = \scmdo(\mathbf{a}^{\prime})$ and $\eta =\scmdo(\mathbf{b}), \omega(\eta) = \scmdo(\mathbf{b}^{\prime})$, such that $\iota \preceq_{\intervsetbase_q} \eta$. Let also the corresponding plans $P^{\iota} \preceq P^{\eta}$ be defined over the empirical measures $\widehat{\prob}_{\scmbase_{\iota}}, \widehat{\prob}_{\scmbase_{\omega(\iota)}}$ and $\widehat{\prob}_{\scmbase_{\eta}}, \widehat{\prob}_{\scmabst_{\omega(\eta)}}$ respectively. We will show now show how a causal constraint may be derived by equating the relationships given by \ot{} and \emph{do-calculus}.
\paragraph{\ot{} relationship.} 
The mass conservation constraints $\mathcal{U}(\pi_{\iota})$ on $P^{\iota}$ induced by \ot{} guarantee that:
\begin{equation}\label{eq:ot_constraints}
    \overbrace{\widehat{\prob}_{\scmbase_{\iota}}(X_j) =  \left ( \sum\nolimits_{i} P^{\iota}_{i,j} \right )_{j}}^{\text{Base}} ~~\forall j \in \dom{\mathbf{X}} \qquad \overbrace{\widehat{\prob}_{\scmabst_{\omega(\iota)}}(X^{\prime}_i ) = \left (\sum\nolimits_{j} P^{\iota}_{i,j}\right )_{i}}^{\text{Abstracted}} ~~\forall i \in  \dom{\mathbf{X^{\prime}}}
\end{equation}

\paragraph{\emph{do-calculus} relationship.}  
Causal inference theory \citep[Chapter 3, pp. 73]{pearl2009causality} relates interventional distributions via the \emph{truncated factorization} (or g-formula). Without loss of generality, let $\pi_{\iota}$ be the pair of observational distributions, where $\iota, \omega(\iota)$ are the null interventions.
Then, it holds that: 
\begin{align}\label{eq:pearl_base}
 \begin{aligned}
\prob_{\scmbase_{\scmdo(\mathbf{b})}}(\mathbf{X})  &= 
 \begin{cases}
\frac{\prob_{\scmbase}(\mathbf{X}) }{ \prod_{i} \prob_{\scmbase}( \mathbf{B}_i = \mathbf{b}_i  ~ \mid ~ \operatorname{PA}(\mathbf{B}_i) ) } &  \text{if } \cmpt{\mathbf{x}}{\scmdo(\mathbf{b})} \\  
 0 &  \text{otherwise}
 \end{cases}
  \end{aligned}   ~~~~ \Bigg\}\text{ Base}
 \end{align}
In our empirical setup, we express \cref{eq:pearl_base} through the minimization of a statistical divergence $d:  \mathbb{R}^D\times  \mathbb{R}^D \to \mathbb{R}_{\geq 0}$, where $D$ is $\cardinality{\dom{\mathbf{X}}}$ for the base and $ \cardinality{\dom{\mathbf{X}^\prime}}$ for the abstracted model, as follows:
\begin{equation}\label{eq:approx_docalc}
    d \left (\widehat{\prob}_{\scmbase_{\scmdo(\mathbf{b})}}(\mathbf{X}) , ~  \frac{1}{\prod_{i} \widehat{\prob}_{\scmbase}( \mathbf{B}_i = \mathbf{b}_i  ~ \mid ~ \operatorname{PA}(\mathbf{B}_i) ) } \widehat{\prob}_{\scmbase}(\mathbf{X}) \right ) \quad\text{ if }\cmpt{\mathbf{x}}{\scmdo(\mathbf{b})} ~~~~ \Bigg\}\text{ Base}
\end{equation}
Throughout, we will work with the general class of Bregman divergences \citep{dhillon2008matrix}. An equivalent relationship to \cref{eq:approx_docalc} for the distributions of the abstracted model may be derived.
\paragraph{Integrating \emph{do-calculus} and \ot{}.} 
Finally, in order to express \cref{eq:approx_docalc} in terms of the optimization variables $P^{\iota}, P^{\eta}$, we substitute in the mass conservation constraints for both the base and the abstracted models given by \cref{eq:ot_constraints}:
\begin{equation}\label{eq:delta_base}
\delta_{\iota, \eta}(P^{\iota}, P^{\eta}) : = d \left( \Big( \sum_{i} P^{\eta}_{i,j} \Big)_j, ~  \frac{1}{(\mathcal{Z}^{\eta})_j} \Big(\sum_i P^{\iota}_{i,j}\Big)_j\right) \quad\text{ if }\cmpt{x_j}{\eta}.  ~ \Bigg\}\text{Base} 
\end{equation}
\begin{equation}\label{eq:delta_abst}
    \delta^{\prime}_{\iota, \eta}(P^{\iota}, P^{\eta}) : = d\left( \Big( \sum_{j} P^{\eta}_{i,j} \Big)_i, ~  \frac{1}{(\mathcal{Z}^{\omega(\eta)})_i} \Big(\sum_j P^{\iota}_{i,j}\Big)_i\right) \quad\text{ if }\cmpt{x_i^{\prime}}{\omega(\eta)}.  ~\Bigg\}\text{Abstracted}
\end{equation}
where $\mathcal{Z}^{\eta}, \mathcal{Z}^{\omega(\eta)}$ are the normalizing vectors for the base and the abstracted distributions respectively, induced from \cref{eq:pearl_base}; see \cref{sec:norm_vectors} for their derivation in terms of the plans. Instead of computing independently the \ot{} plans as in \cref{eq:optproblem} we can jointly learn plans that preserve causal relations by incorporating the base and abstracted model distances $\mathcal{D}(P^{\iota}, P^{\eta})=[\delta_{\iota, \eta},~ \delta^{\prime}_{\iota, \eta}]^\top$ defined over the marginals of two plans.

\subsection{A causally informed cost function}\label{sec:omegacost}
The \ot{} cost function captures the transport problem's geometry in order to find the optimal plan. Although in $\mathbb{R}$ costs like $l_p$ can represent the distance between samples, this is not trivial when dealing with samples between causal models. However, the $\omega$ map of the $\tauomega$ transformation provides a solution by formally encoding the interventional relationship between samples of two \scm{s}. In order to compute a distance between samples $\mathbf{x} \in \dom{\mathbf{X}}$ of the base and $\mathbf{x}^{\prime} \in \dom{\mathbf{X}^{\prime}}$ of the abstracted model, given interventions $\iota = \scmdo(\mathbf{a})$ and $\omega(\iota) = \scmdo(\mathbf{a}^{\prime})$, we exploit $\omega$ to discount the cost of transporting sample $\mathbf{a}$ to $\mathbf{a}^{\prime}$. We then define $c_{\omega}: \dom{\mathbf{X}} \times \dom{\mathbf{X}^{\prime}} \to \mathbb{R}_{\geq 0}$:
\begin{align}\label{eq:omega_cost_func}
    c_{\displaystyle \omega}(\mathbf{x},\mathbf{x}^\prime) = \cardinality{\intervsetbase} - \sum_{\iota \in \intervsetbase} 
    \indicator \left[ \cmpt{\mathbf{x}}{\iota} \logicaland \cmpt{\mathbf{x}^\prime}{\omega(\iota)}\right],
\end{align}
where $\indicator[a]$ is the indicator function returning one if the condition $a$ is satisfied. The function $c_{ \omega}$ discounts the cost of transporting the sample $\mathbf{x}$ to $\mathbf{x^{\prime}}$ proportionally to the number of pairs $(\iota,\omega(\iota))$ w.r.t. which $\mathbf{x}$ and $\mathbf{x^{\prime}}$ are compatible. Hence, the larger and more diverse the set of pairs $\intervpairs(\intervsetbase)$ is, the more informative the $\omega$-cost will be, thereby enhancing its capacity to convey comprehensive insights into the cost matrix. This sensitivity of  $c_{\omega}$ to $\intervsetbase$ is demonstrated in \cref{sec:experiments}. Finally, by construction, $C_{\omega}$ has the advantage of being is invariant to the ordering of the values $\mathbf{x}$ (columns) and $\mathbf{x}^{\prime}$ (rows). In \cref{sec:costs} we offer an illustration of a $C_{\omega}$ matrix derived from $c_{\omega}$ and further discuss the construction of $\omega$-costs.

\subsection{The causal optimal transport of abstractions (\cota{}) objective}\label{sec:cota}
We now discuss how we can integrate the \emph{do-calculus} constraints discussed in \cref{sec:constraints} and the $\omega$-informed cost presented in \cref{sec:omegacost} in the the \ot{} framework to jointly solve multiple transport problems and learn an abstraction $\tau$. For a given set of pairs $\intervpairs(\intervsetbase_{k}) = \left\{\pi_{\iota_1},\cdots	\pi_{\iota_N}\ \mid \iota_n \in \intervsetbase_{k}\right\}$ where $\intervsetbase_{k}$ a maximal chain, we define the objective function of \cota{} as the following \ot{} problem:
\begin{align}\label{eq:objective_2d}
 \resizebox{.99\hsize}{!}{$ \bm{P}^{\star}_{k} = \cota_c\left(\intervpairs\left(\intervsetbase_{k}\right)\right) = \displaystyle \argmin_{\substack { \{ P^{\iota_n} \in \mathcal{U}(\pi_{\iota_n})} \}_{\iota_n \in \intervsetbase_{k}} } \left\{\kappa \cdot \sum_{\iota_n \in \intervsetbase_{k}}\underbrace{\left<C_\omega,P^{\iota_n}\right>}_\text{\ot{}} +\bm{\lambda}^\top   \underbrace{\mathcal{D}(P^{\iota_n}, P^{\iota_{n+1}})}_\text{do-calculus constraints} - \mu \cdot \underbrace{\mathcal{H}(P^{\iota_n}) }_\text{entropy}\right\} , $}
\end{align} 
where $\bm{\lambda} = \left[ \lambda, \lambda^{\prime}\right]^\top$ and $(\kappa, \bm{\lambda}, \mu)$ a convex combination, i.e. $\kappa+ \lambda+\lambda^{\prime}+\mu = 1$ for $\kappa, \lambda, \lambda^{\prime}, \mu \geq 0$. Thus, we transformed the initial \ca{} problem of \cref{eq:optproblem} into a joint multi-marginal \ot{} problem integrated with causal knowledge from different sources. \hyperref[sec:cota_alg]{Algorithm 1} presents the end-to end implementation of \cota{}. The complexity of the algorithm is $\mathcal{O}(N_{\max} + N_{\text{chains} } \cdot (D_{\max}^3 \log D_{\max}) )$, where $N_{\max} = \max{(N, N^{\prime})}$ with $N, N^{\prime}$ respectively the number of samples for base and abstracted model, $N_{\text{chains}}$ the number of maximal chains, and $D_{\max} = \max{(D , D^{\prime})}$ with $D = \cardinality{\dom{\mathbf{X}}}, D^{\prime} = \cardinality{\dom{\mathbf{X}^\prime}} $. The first term accounts for the complexity of line 3, while the second term accounts for the loop of line 13 and the internal call of the \cota{} solver. The entropy regularization $\mathcal{H}(\mathcal{P}^{\iota_n})$ allows for a speed up to $\mathcal{O}(D^2)$ with Sinkhorn algorithms \citep{peyre2019computational}. In \cref{sec:approx_cota} we also present \texttt{Approximate} \cota{}, an approximation that halves the causal constraints $\mathcal{D}(\cdot, \cdot)$ of the optimisation problem of \cref{eq:objective_2d}. 
\begin{algorithm}[htb]
    \caption{\cota{}}
    \label{sec:cota_alg}
    \begin{algorithmic}[1]
    \REQUIRE \dirag{}s for $\scmbase[\mathbf{X}], \scmabst[\mathbf{X^{\prime}}]$, sets $\intervsetbase, \intervsetabst$ and  $\omega:\intervsetbase \to \intervsetabst$, surjective and order-preserving.
    \ENSURE $\tau:\dom{\mathbf{X}} \to \mathcal{A}^{|\dom{\mathbf{X^{\prime}}}|}$, where $\mathcal{A}^n = \left\{\bm{p} \in \mathbb{R}^n:~ p_i \geq 0, \sum_i p_i =1 \right\}$.\vspace{1.1mm}
    \STATE \textbf{for }{$\iota \in \intervsetbase$} \textbf{do:}
    \STATE ~~~~$\{ \mathbf{x}_{j} \}_{j=1}^{N} \sim 
    \prob_{\scmbase_{\iota}}(\mathbf{X})$, $\{ \mathbf{x}_{i}^{\prime} \}_{i=1}^{M} \sim \prob_{\scmabst_{\omega(\iota)}}(\mathbf{X}^{\prime})$ ~~~ \textit{\# sampling from the true distributions}\vspace{1.1mm}
    \STATE ~~~~ $\widehat{\prob}_{\scmbase_{\iota}}(\mathbf{X}) \leftarrow \sum_{j=1}^{N} \alpha_j \delta_{\mathbf{x}_{j}}$, $ ~\widehat{\prob}_{\scmabst_{\omega(\iota)}}(\mathbf{X^{\prime}}) \leftarrow \sum_{i=1}^{M} \beta_i \delta_{\mathbf{x}_{i}^{\prime}}$ ~ \textit{\# construct the empirical measures}\vspace{1.1mm}
   \STATE \textbf{for $j=0$ to $N$} \textbf{do:}
    \STATE ~~~~\textbf{for $i=0$ to $M$} \textbf{do:}
    \STATE ~~~~~~~~$C_{\omega}[i,j] \leftarrow \cardinality{\intervsetbase} - \sum\nolimits_{\iota \in \intervsetbase} 
    \indicator \left[ \cmpt{\mathbf{x_j}}{\iota} \logicaland \cmpt{\mathbf{x_i}^\prime}{\omega(\iota)}\right]$ \textit{\# compute the $\omega$-cost matrix}\vspace{1.1mm}
     \STATE $\mathcal{C}(\intervsetbase) \leftarrow \texttt{compute\_chains}(\intervsetbase)$ ~~~ \textit{\# compute the set of all maximal chains of $\intervsetbase$}\vspace{1.1mm}
     \STATE \textbf{for }{$\intervsetbase_k \in \mathcal{C}(\intervsetbase)$} \textbf{do:}
     \STATE ~~~~$\intervpairs(\intervsetbase_k) \leftarrow \emptyset$
    \STATE ~~~~\textbf{for }{$\iota \in \intervsetbase_k$} \textbf{do:}
     \STATE ~~~~~~~~$\intervpairs(\intervsetbase_k) \leftarrow \intervpairs(\intervsetbase_k) \cup \{(\widehat{\prob}_{\scmbase_{\iota}}, \widehat{\prob}_{\scmabst_{\omega(\iota)}})\}$  \textit{\# compute the set of pairs for every $\intervsetbase_k$}\vspace{1.1mm}
     \STATE $\mathcal{P}\leftarrow \emptyset$
     \STATE \textbf{for }{$\intervsetbase_k \in \mathcal{C}(\intervsetbase)$} \textbf{do:}
     \STATE ~~~~$\mathcal{P} \leftarrow \mathcal{P}\cup\{\cota_c(\intervpairs\left(\intervsetbase_{k}\right)\}$ ~~\textit{\# run \cota{}  for every $\intervsetbase_k$ and assemble the set of all plans}\vspace{1.1mm}
     \STATE \textbf{if} \cota{$(\widehat{\mathcal{P}})$} \textbf{ then } $\widehat{\mathcal{P}} \leftarrow \frac{1}{\vert \mathcal{P}\vert}\sum\nolimits_{t=1}^{\vert \mathcal{P}\vert} P_{t}$
     \STATE \textbf{~~~~Return: }$\tau \leftarrow  f_s(\widehat{\mathcal{P}})$ ~~~\textit{\# return the final $\tau$ as the stochastic mapping $f_s(\cdot)$ of the average plan }\vspace{1.1mm}
     \STATE \textbf{if} \cota{$(\widehat{\tau})$} \textbf{ then } $\widehat{f_s(P_{t})} \leftarrow \frac{1}{\vert \mathcal{P}\vert}\sum\nolimits_{t=1}^{\vert \mathcal{P}\vert} f_s(P_{t})$
     \STATE \textbf{~~~~Return: }$ \tau \leftarrow \widehat{f_s(P_{t})}$ ~\textit{\# return the final $\tau$ as the average stochastic mapping $f_s(\cdot)$ of the plans}
    \end{algorithmic}
\end{algorithm}

\begin{theorem*}[Convexity of \cota{}]\label{def:convthm}
The optimization problem given by \cref{eq:objective_2d} is jointly convex in the transport plans $ \{ P^{\iota_n} \in \mathcal{U}(\pi_{\iota_n}) \}_{\iota_{n} \in \intervsetbase_{k}} ~ \forall \intervsetbase_{k} \in \mathcal{C}(\intervsetbase)$. When $\mu > 0$, it is strictly convex.
\end{theorem*}

\textbf{Proof (sketch)}
The main assertion that needs to be shown is the joint convexity of the function $ \{ P^{\iota_n} \}_{\iota_{n} \in \intervsetbase_{k}} \to \kappa \cdot \sum_{\iota_n \in \intervsetbase_{k}} \left<C_\omega,P^{\iota_n}\right> + \bm{\lambda}^\top \cdot \mathcal{D}(P^{\iota_n}, P^{\iota_{n+1}}) + \mu \cdot \mathcal{H}(P^{\iota_n}) $ in all of the plans. \footnote{Other conditions, like convex constraints and domain properties, are satisfied by stochastic matrices; see \cref{sec:proof2}.} To show the main assertion, we use two inequalities following directly from the convexity definition of the single plan function $P^{\iota_n} \to  \left<C_\omega,P^{\iota_n}\right>$ and from the joint convexity of $( P^{\iota_n}, P^{\iota_{n+1}}) \to \mathcal{D}(P^{\iota_n}, P^{\iota_{n+1}}) $ in $P^{\iota_{n}}, P^{\iota_{n+1}}$. Further, note that the entropic regulariser $\mathcal{H}(P^{\iota_n}) $ is known to be a strictly convex function of $P^{\iota_n}$ \citep{peyre2019computational}. Finally, since the function defined by the set of all the plans in $\intervsetbase_{k}$ is a summation, combining these two inequalities gives the desired result, with strict convexity holding for $\mu > 0$ due to $\mathcal{H}(P^{\iota_n})$. The full proof is provided in \cref{sec:proof2}.

\section{Related work}
In this section we briefly review related works of \ot{} within the domain of causality and the multi-marginal techniques akin to our own methodology. There has been increasing interest in applying \ot{} methodology to perform inference in causal models. Regarding treatment effect estimation, \citet{torous2021optimal} propose estimators of binary treatment effects in the potential outcome framework based on \ot{} to handle high-dimensional covariates; \citet{gunsilius2021matching} address covariate matching in multi-valued treatments via multi-marginal \ot{}. Recently, \citet{tu2022optimal} propose the use of \ot{} to perform bivariate causal discovery in the context of continuous data and additive noise, with benefits such as avoiding to specify likelihoods and efficient computation due to one-dimensional distributions. Additionally, an extension of \ot{} to multiple marginals is considered in \citet{peyre2019computational,pass2015multi,kostic2022batch}. In this setting, the problem involves finding couplings between source and target distributions, even in high-dimensional cases. Differently from our formulation, this multi-marginal setup does not consider any relations between the computed transport plans.

\section{Experiments}\label{sec:experiments}
Throughout the experiments, we investigate the performance of \cota{} under diverse experimental settings and in different tasks in order to showcase: \textbf{(a)} its superiority over non-causal solutions; \textbf{(b)} the actual gains of introducing the \emph{do-calculus} constraints into the optimization routine; \textbf{(c)} the advantage of the causally informed $\omega$-cost, relative to the size and diversity of the intervention set, compared to standard/non-causal costs; and \textbf{(d)} the efficiency of \cota{} as a data augmentation tool in a downstream task compared to established state-of-the-art \ca{} learning frameworks. The code and results for all experiments are publicly accessible \footnote{\href{https://github.com/yfelekis/COTA}{github.com/yfelekis/COTA}}.

\paragraph{\textbf{\cota{}.}} We run \cota{} considering two Bregman divergences for the distance term $\mathcal{D}$ of \cref{eq:objective_2d}, the {Frobenius norm} (\fro) and the {Jensen–Shannon Divergence} (\jsd). We also run an ablation study in which we replace $c_{\omega}$ in \cota{'s} objective with a conventional cost $c_{\mathcal{H}}$ based on the Hamming distance and compare the two costs' performance. Finally, regarding the parameter $\bm{\lambda}=\left[ \lambda, \lambda^{\prime} \right]^\top$ in \cref{eq:objective_2d}, in the main paper we demonstrate the equal weight case where $\lambda =\lambda^{\prime}$\footnote{i.e. $\bm{\lambda}^\top\cdot\mathcal{D}(P^{\iota}, P^{\eta}) = [\lambda, \lambda]\cdot \begin{bmatrix}
           \delta_{\iota, \eta}  \\
           \delta^{\prime}_{\iota, \eta}
         \end{bmatrix}= \lambda\cdot(\delta_{\iota, \eta}+\delta_{\iota, \eta}^{\prime})$.} and provide additional results from the more general case of $\lambda \neq \lambda^{\prime}$ in \cref{sec:complete_results}.

\paragraph{Baselines.} In addition, we compare our method with three alternative setups which serve as baselines. These configurations consist of non-causal independent solutions of the \ot{} problem, as described in \cref{eq:optproblem}, or barycentric adaptations of the standard \ot{} framework. In particular: 
\begin{itemize}[itemsep=0.01pt]
    \item In \pairOT{} we apply $\ot_c$ to generate a set of $k$ independent plans $\mathcal{P} = \{P_1,\ldots,P_k\}$ and aggregate them into an average single plan $\widehat{\mathcal{P}}$ and compute $\tau= f_s(\widehat{\mathcal{P}})$. This is equivalent to \cota{$(\widehat{\mathcal{P}})$} when $\kappa =1$, $\lambda=0$ and $\mu=0$. 
    \item In \baryOT{} we first compute two barycenters: one of the base model's  ($\bar{\bm{\alpha}}$) and one of the abstracted model's distributions ($\bar{\bm{\beta}}$) and then solve the standard $\ot_c$ (single-pair) problem for $(\bar{\bm{\alpha}}, \bar{\bm{\beta}})$, to compute the plan $\bar{P}$, and finally compute $\tau = f_s(\bar{P})$.
    \item In \mapOT{} we apply $\ot_c$ to generate the set of $k$ plans $\mathcal{P} = \{P_1,\ldots,P_k\}$, compute the $k$ independent stochastic maps from them and compute $\tau$ as the average of them $\tau = \widehat{\left\{f_s(P_i)\right\}}_{i \in [K]}$. This is equivalent to \cota{$(\widehat{\tau})$} when $\kappa =1$, $\lambda=0$ and $\mu=0$.
\end{itemize}

\paragraph{Evaluation.} Across all scenarios, we assess the learned $\tau$ in terms of the abstraction error from \cref{eq:abst_error} using the \jsd{} ($e_{\jsd}(\tau)$) and the Wasserstein ($e_{\wass}(\tau)$) distances by employing a leave-one-pair-out procedure to measure the quality and the robustness of the learned abstraction. Specifically, we remove one pair $\pi_i=(\alpha_i,\beta_i)$ from $\intervpairs(\intervsetbase)$, learn $\tau$ from the remaining pairs, and measure the $e(\tau)$ distance between ${\tau}_{\#}(\alpha_i)$ and $\beta_i$. All reported measures are presented as the mean and standard deviation over 50 repetitions with a 95\% confidence interval. For \cota{} we select the hyperparameters $(\kappa, \lambda, \mu)$ via a grid-search of 100 convex combinations. 

\subsection{Causal Abstraction Learning Simulations}
The \dirag{s} alongside their intervention sets for all the following scenarios are presented in \cref{sec:dags_chains}.

\paragraph{Simple Lung Cancer (STC).} This motivating example is made up by a discrete base model with a chain structure ($\operatorname{Smoking} \to \operatorname{Tar} \to \operatorname{Cancer}$) and an abstracted that removes the mediator node. We investigate two different scenarios: when the interventions are performed on variables \textbf{(a)} without parents (\textbf{STC}$_{\text{np}}$) and \textbf{(b)} with parents (\textbf{STC}$_{\text{p}}$). \autoref{tab:STC_NP_results_main} showcases the abstraction error of \cota{} in \textbf{STC}$_{\text{np}}$ and demonstrates its superiority against all the different baselines, both for $e_{\wass}(\tau)$ and $e_{\jsd}(\tau)$. Also, notice how $c_{\omega}$ returns a lower abstraction error compared to $c_{\mathcal{H}}$, not only in different settings of \cota{}, but also with baseline ones. This suggests that indeed the $\omega$-cost can provide more relevant information for learning an abstraction. Further, in \cref{fig:ternaryplot} we highlight the impact of the parameter $\lambda$, which weighs the \emph{do-calculus} constraints' term in \cref{eq:objective_2d}; as the best performing setting of \cota{} is reached for $\lambda >0$, this validates the hypothesis that introducing causal constraints helps learning better abstractions. In the \textbf{STC}$_{\text{p}}$ scenario, \cota{} still outperforms the baselines (see \cref{sec:complete_results}), although \autoref{tab:STC_P_results_main} shows that $c_{\mathcal{H}}$ returns a lower abstraction error compared to $c_{\omega}$. We argue that this is likely due to the dependency of $\omega$-cost on the intervention set, which, in this case, comprises only two interventions. As explained in \cref{sec:omegacost} such a small intervention set leads to an almost-uniform and uninformative $\omega$-cost, and in such case a conventional cost like Hamming might be able to capture certain patterns more efficiently. Visualizations of the induced cost matrices for both functions can be found in \cref{fig:omega_costs} and \cref{fig:ham_costs} in \cref{sec:costs}.
\begin{table}[t]
\begin{minipage}[b]{0.49\linewidth}
\centering
\label{tab:syntexp_b_performance}
\resizebox{\columnwidth}{!}{%
\begin{tabular}{ccccc}
    \toprule
    Method & $\mathcal{D}$  & $\mathcal{C}$ & $e_{\jsd}(\tau)$   & $e_{\wass}(\tau)$\\
    \midrule
     \cota{($\widehat{P}$)} & \fro  & $c_{\omega}$ & $\bm{0.010 \pm 0.005}$&$\bm{0.011 \pm 0.003}$\\
     &  & $c_{\mathcal{H}}$ & $0.087 \pm 0.007$&$0.025 \pm 0.001$\\
     \\
     & \jsd  & $c_{\omega}$ & $0.012 \pm 0.006$&$0.012 \pm 0.003$\\
     &   & $c_{\mathcal{H}}$ & $0.087 \pm 0.006$ &$0.025 \pm 0.001$\\
     \\
    \midrule
     \cota{($\widehat{\tau}$)} & \fro  & $c_{\omega}$ & $0.013 \pm 0.021$ &$0.171 \pm 0.001$\\
     &  & $c_{\mathcal{H}}$ & $0.169 \pm 0.005$ &$0.178 \pm 0.001$\\
     \\
     & \jsd  & $c_{\omega}$ & $0.014 \pm 0.021$&$0.171 \pm 0.001$\\
     &   & $c_{\mathcal{H}}$ & $0.169 \pm 0.004$ &$0.178 \pm 0.001$\\
     \\
    \midrule
    \pairOT & --   & $c_{\omega}$    & $0.013 \pm 0.002$ &$0.011 \pm 0.002$\\
    & -- &    $c_{\mathcal{H}}$  & $0.093 \pm 0.004$ &$0.039 \pm 0.002$\\
    \midrule
    \mapOT{} & -- &  $c_{\omega}$    & $0.023 \pm 0.022$ &$0.147 \pm 0.001$\\
           & -- &  $c_{\mathcal{H}}$  & $0.169 \pm 0.022$ &$0.156 \pm 0.001$\\
    \midrule
    \baryOT{} & -- &  $c_{\omega}$    & $0.233 \pm 0.142$ &$0.067 \pm 0.042$\\
              & -- &  $c_{\mathcal{H}}$  & $0.323 \pm 0.074$ &$0.095 \pm 0.039$\\
    \bottomrule
\end{tabular}}
\captionsetup{skip=10pt}
\caption{{Abstraction error evaluation for the \textbf{STC}$_{\text{np}}$ example. The configuration $\cota{(\widehat{P})}-\fro - c_{\omega}$ yields the lowest abstraction error when compared to all other settings and baselines, for both metrics. A "rich" intervention set $c_{\omega}$ effectively captures the correspondence between samples leading to superior performance over $c_{\mathcal{H}}$ in all \cota{} settings and baselines.}}
\label{tab:STC_NP_results_main}
\end{minipage}\hfill%
\begin{minipage}[b]{0.49\linewidth}
\resizebox{\columnwidth}{!}{%
\includegraphics{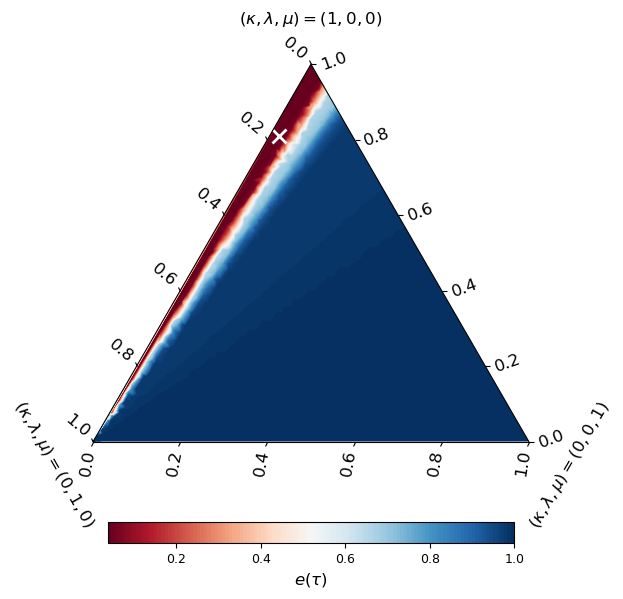}}
\captionsetup{skip=10pt}
\captionof{figure}{{Effect of $\lambda$ for the \textbf{STC}$_{\text{np}}$ example. The ternary plot illustrates a grid-search amongst 1000 convex combinations of $(\kappa, \lambda, \mu)$ for the $\cota{(\widehat{P})} - \fro - c_{\omega}$ setting. Values of $\lambda$ close to zero present high abstraction error, demonstrating the benefit of the \emph{do-calculus} constraints in the \ot{} problem. The minimum is reached at (.81, .17, .02) and is denoted with "$\bm{\times}$".}}
\label{fig:ternaryplot}
\end{minipage}
\end{table}

\begin{table}[t]
\begin{minipage}[b]{0.49\linewidth}
\resizebox{\columnwidth}{!}{%
\begin{tabular}{ccccc}
    \toprule
    Method & $\mathcal{D}$  & $\mathcal{C}$ & $e_{\jsd}(\tau)$   & $e_{\wass}(\tau)$\\
    \midrule
    \cota{($\widehat{\tau}$)} & \fro  & $c_{\omega}$ & $0.249\pm 0.005$ &$0.135 \pm 0.001$\\
     &  & $c_{\mathcal{H}}$ & $\bm{0.229\pm 0.003}$&$0.129 \pm 0.001$ \\
    \bottomrule
        \end{tabular}}
        \captionsetup{skip=10pt}
\caption{{Abstraction error for the \textbf{STC}$_{\text{p}}$ example with $\intervsetbase = \{\scmdo(\operatorname{T}=0), \scmdo(\operatorname{T}=1)\}$. Cost $c_{\mathcal{H}}$ outperforms $c_{\omega}$ due to limited intervention set $\intervsetbase$.}}
\label{tab:STC_P_results_main}
\end{minipage}\hfill%
\begin{minipage}[b]{0.49\linewidth}
\resizebox{\columnwidth}{!}{%
\setlength{\tabcolsep}{10pt} 
\renewcommand{\arraystretch}{.925}
\begin{tabular}{ccccc}
    \toprule
    Method & $\mathcal{D}$  & $\mathcal{C}$ & $e_{\jsd}(\tau)$   & $e_{\wass}(\tau)$\\
    \midrule
      \cota{($\widehat{\tau}$)} & \fro  & $c_{\omega}$ & $0.379$ &$0.053 $\\
     &  & $c_{\mathcal{H}}$ & $\bm{0.220}$&$0.053 $ \\
    \bottomrule
        \end{tabular}}
        \captionsetup{skip=10pt}
\caption{Abstraction error for the \textbf{EBM} example. The conventional cost $c_{\mathcal{H}}$ outperforms $c_{\omega}$ due to a limited intervention set $\intervsetbase$.}
\label{tab:EBM_results_main}
\end{minipage}
\end{table}

\paragraph{LUng CAncer Set (LUCAS)}
This model\footnote{\href{http://www.causality.inf.ethz.ch/data/LUCAS.html}{http://www.causality.inf.ethz.ch/data/LUCAS.html}} is a large-scale synthetic \scm{} designed to simulate data related to the study of lung cancer. \autoref{tab:lucas_results_main} confirms that even on larger and more realistic problems \cota{} exceeds the baselines in terms of the abstraction error, and $c_{\omega}$ provides results at least as good as $c_{\mathcal{H}}$. The full table of results can be found in \cref{sec:complete_results}.

\paragraph{Electric Battery Manufacturing (EBM)}\label{sec:ebm} 
Finally we compare with the only real-world public dataset that, to the best of our knowledge, has been used for \ca{} learning \citep{zennaro2023jointly}. This dataset contains data related to electric battery manufacturing collected from two experimental settings. The first setting (WMG) has been modelled through a low-level \scm{} that captures the effect of a control variable (\emph{comma gap}) on an output (\emph{mass loading}) at multiple spatial locations. The second setting (LRCS) is modelled through a high-level \scm{} that relates the same control variable to a single output \citep{cunha2020artificial}. We learn an abstraction map $\tau$ using the set of real interventions performed during the collection of the data. Following \cite{zennaro2023jointly}, we use the learned map $\tau$ to abstract the WMG data and aggregate them with the LRCS data; then we perform a set of downstream regression tasks. Compared to the SOTA which required full knowledge of the underlying \scm{}, \cota{} requires only knowledge of the underlying \dirag{} and provides better results in terms of the Mean Square Error (MSE) in all the proposed setups, as shown in \autoref{tab:downstream}. In general, \cota{} is always the top performer compared to the baselines (see \cref{sec:complete_results} for the complete table of results), while in \autoref{tab:EBM_results_main} we highlight a similar behaviour as the one in \autoref{tab:STC_P_results_main} whereby the limited size of the interventional data prevents $c_{\mathcal{H}}$ to lead to better abstractions compared to $c_{\omega}$. A complete presentation of the different settings and the data of this case study can be found in \cref{sec:downstream_details}.
\begin{table}[t]
\begin{minipage}[t]{0.48\linewidth}
\resizebox{\columnwidth}{!}{%

\begin{tabular}{ccccc}
    \toprule
    Method & $\mathcal{D}$  & $\mathcal{C}$ & $e_{\jsd}(\tau)$   & $e_{\wass}(\tau)$\\
    \midrule
    \cota{($\widehat{P}$)} & \fro  & $c_{\omega}$ & $0.287 \pm 0.014$&$\bm{0.044 \pm 0.001}$\\
     &  & $c_{\mathcal{H}}$ & $0.287 \pm 0.014$&$0.047 \pm 0.001$\\
     \midrule
     \cota{($\widehat{\tau}$)} & \fro  & $c_{\omega}$ & $\bm{0.263 \pm 0.005}$ &$0.061 \pm 0.001$\\
     &  & $c_{\mathcal{H}}$ & $\bm{0.263 \pm 0.006}$ &$0.061 \pm 0.001$\\
    \midrule
    \pairOT & --   & $c_{\omega}$    & $0.306 \pm 0.009$ &$0.045 \pm 0.001$\\
    & -- &    $c_{\mathcal{H}}$  & $0.387 \pm 0.002$ &$0.047 \pm 0.001$\\
    \midrule
    \mapOT{} & -- &  $c_{\omega}$    & $0.294 \pm 0.008$ &$0.054 \pm 0.001$\\
           & -- &  $c_{\mathcal{H}}$  & $0.350 \pm 0.005$ &$0.054 \pm 0.001$\\
    \midrule
    \baryOT{} & -- &  $c_{\omega}$    & $0.294 \pm 0.047$ &$0.044 \pm 0.003$\\
              & -- &  $c_{\mathcal{H}}$  & $0.414 \pm 0.040$ &$0.046 \pm 0.010$\\
    \bottomrule
        \end{tabular}}
        \captionsetup{skip=10pt}
\caption{{Abstraction error for \textbf{LUCAS}. The configuration $\cota{(\widehat{\tau})}-\fro$ yields the lowest abstraction error for the \jsd{} metric and the $\cota{(\widehat{P})}-\fro - c_{\omega}$ setting for the \wass{} metric. The $\omega$-cost $c_{\omega}$ outperforms $c_{\mathcal{H}}$ .}}
\label{tab:lucas_results_main}
\end{minipage}\hfill%
\begin{minipage}[t]{0.5\linewidth}
\resizebox{\columnwidth}{!}{%
    \setlength{\tabcolsep}{10pt} 
\renewcommand{\arraystretch}{3.475}
  \begin{tabular}{cccc}
    \toprule
    \Large Training set & \Large Test set & \Large \cite{zennaro2023jointly} & \Large \cota{}\\
    \midrule \\
   \Large LRCS[$CG\neq k$] & \Large LRCS[$CG = k$] & \huge $1.86 \pm 1.75$ & \huge \bm{$1.40 \pm 1.39$} \\ 
    \midrule
  \Large LRCS[$CG\neq k$] & \Large LRCS[$CG = k$] &\huge $0.22 \pm 0.26$ &\huge \bm{$0.19 \pm 0.04$ }\\
   \Large +WMG & & \\  
    \midrule
    \Large LRCS[$CG\neq k$] & \Large LRCS[$CG = k$] & \huge $1.22 \pm 0.95$ &\huge \bm{ $0.80 \pm 0.55$ }\\ 
     \Large +WMG[$CG\neq k$] & \Large WMG[$CG = k$] &  \\
    \bottomrule
  \end{tabular}}
  \captionsetup{skip=10pt}
\caption{{MSE of \cota{} and a SOTA \ca{} framework on a regression task for \textbf{EBM}. Augmenting
data via the learned abstraction reduces the average error in all different settings compared to the SOTA. We used $\cota{(\widehat{P})}-\fro - c_{\omega}$ with the hyperparameters $(\kappa, \lambda, \mu) = (.2, .5, .3)$ achieving the lowest abstraction error.}}
\label{tab:downstream}
\end{minipage}
\end{table}

\section{Discussion}
In this work, we presented \cota{}, a framework for learning causal abstractions from observational and interventional data through a causally constrained multi-marginal \ot{} formulation. Incorporating \emph{do-calculus} constraints and a causally-informed cost $c_{\omega}$ in the optimisation problem led to lower abstraction error compared to non-causal baselines in diverse scenarios. The effectiveness of the $c_{\omega}$ cost was shown to be sensitive to the intervention set; expanding the interventions set improved the performance with the $\omega$-cost compared to conventional non-causal costs like Hamming. Lastly, \cota{} outperform the prior \ca{} learning art of \cite{zennaro2023jointly} when employed as a data augmentation procedure on a real world regression task. 

Our work opens up new challenges and directions in both fields of causal abstraction learning and \ot{}. First, constrained multi-marginal \ot{} settings like \cota{} have been understudied in the literature, and further theoretical work on the guarantees of existence and uniqueness of the estimated maps is needed. Recent methods for joint learning of plans and parameterised maps \citep{uscidda2023monge, mapestimation2018flamary} to obtain better estimators \citep{perrot2016mapping} hold promise in this front. Furthermore, generalising a framework like \cota{} to semi-Markovian \scm{s} presents a significant challenge, because lifting the causal sufficiency assumption may render the estimation of certain causal constraints unidentifiable. Finally, another interesting direction would be to extend \ca{} learning frameworks like \cota{} in order to incorporate temporal dependencies and continuous-time models, for example structural dynamical causal models \citep{bongers2018causal}.

\section*{Acknowledgments}
\textbf{YF}: This scientific paper was supported by the Onassis Foundation - Scholarship ID: F ZR 063-1/2021-2022. \textbf{TD}: Acknowledges support from a UKRI Turing AI acceleration Fellowship [EP/V02678X/1]. The authors would also like to acknowledge the University of Warwick Research Technology Platform (RTP) for assistance in the research described in this paper and the EPSRC platform for ensemble computing "Sulis" [EP/T022108/1].

\clearpage
\section*{Appendix}
\begin{appendix}

\section{Derivation of normalizing vectors}\label{sec:norm_vectors}
We follow the notation introduced in the main text. For $\iota \preceq \eta$,
$\iota  = \scmdo(\mathbf{A}=\mathbf{a})$ and $\omega(\iota) = \scmdo(\mathbf{A}^{\prime}=\mathbf{a}^{\prime})$ induce a transport plan $P^{\iota}$, while $\eta =\scmdo(\mathbf{B}=\mathbf{b})$ and $\omega(\eta) = \scmdo(\mathbf{B}^{\prime}=\mathbf{b}^{\prime})$ induce a transport plan $P^{\eta}$. Our aim is to express the normalizing vectors $ \mathcal{Z}^{\eta} = \mathbb{P}_{\scmbase}( \mathbf{B}_i = \mathbf{b}_i   \mid  \operatorname{PA}(\mathbf{B}_i) )$ and $ \mathcal{Z}^{\omega(\eta)} = \mathbb{P}_{\scmabst}( \mathbf{B^{\prime}}_i = \mathbf{b^{\prime}}_i  \mid \operatorname{PA}(\mathbf{B^{\prime}}_i))$ in terms of the plan $P^{\iota}$. These are the conditional probabilities defined in \cref{eq:pearl_base} and can be written as: 
\begin{equation}\label{eq:norm_vecs_plan_base}
    \mathbb{P}_{\scmbase}( \mathbf{B}_i = \mathbf{b}_i   \mid  \operatorname{PA}(\mathbf{B}_i) ) = \frac{\mathbb{P}_{\scmbase}( \mathbf{B}_i = \mathbf{b}_i,  \operatorname{PA}(\mathbf{B}_i) )}{\mathbb{P}_{\scmbase}( \operatorname{PA}(\mathbf{B}_i) )} ~~ \Bigg\}\text{ Base}
\end{equation}
\begin{equation}\label{eq:norm_vecs_plan_abst}
\mathbb{P}_{\scmabst}( \mathbf{B^{\prime}}_i = \mathbf{b^{\prime}}_i   \mid  \operatorname{PA}(\mathbf{B^{\prime}}_i) ) = \frac{\mathbb{P}_{\scmabst}( \mathbf{B^{\prime}}_i = \mathbf{b^{\prime}}_i,  \operatorname{PA}(\mathbf{B^{\prime}}_i) )}{\mathbb{P}_{\scmabst}( \operatorname{PA}(\mathbf{B^{\prime}}_i) )} ~~ \Bigg\}\text{ Abstracted}
\end{equation}
Then given $P^{\iota} \preceq P^{\eta}$ we express the sub-parts of the equations above in terms of the transportation plan $P^{\iota}$ by defining specific sets of indices on it. Starting from \cref{eq:norm_vecs_plan_base} and the denominator $\mathbb{P}_{\scmbase}( \operatorname{PA}(\mathbf{B}_i) )$, this can be then written as:
\begin{equation}
    \mathbb{P}_{\scmbase}( \operatorname{PA}(\mathbf{B}_i) ) = \sum_{[i,j]\in \mathcal{O}_{\iota, \eta, \rho}}P_{ij}^{\iota},
\end{equation}
where $ \mathcal{O}_{\iota, \eta, \rho} = \left\{ [i,j]~ \mid ~x_j \in \dom{\mathbf{X}} \land \cmpt{x_j}{\scmdo(\operatorname{PA}_{\mathbf{B}}=\rho)}\right\},~ \text{for } \rho \in \dom{\operatorname{PA}_{\mathbf{B}}}$.

Consequently, for the numerator we first define the following set: 
$$\mathcal{C}_{\iota,\eta} = \left\{ [i,j]~ \mid ~x_j \in \dom{\mathbf{X}} \land \cmpt{x_j}{\eta}\right\}$$
and also let the intersection set $\Omega_{\iota,\eta, \rho} = \mathcal{C}_{\iota,\eta} \cap \mathcal{O}_{\iota,\eta, \rho}$ for every $\rho \in \dom{\operatorname{PA}_{\mathbf{B}}}$. Then, we have:
\begin{equation}
    \mathbb{P}_{\scmbase}( \mathbf{B}_i = \mathbf{b}_i,  \operatorname{PA}(\mathbf{B}_i) ) = \sum_{[i,j]\in \Omega_{\iota,\eta, \rho}}P^{\iota}_{i,j}
\end{equation}

By performing the symmetric operations for the abstracted model in \cref{eq:norm_vecs_plan_abst} and get the respective sets $\mathcal{O}^{\prime}_{\iota, \eta, \rho^{\prime}}, \mathcal{C}^{\prime}_{\iota,\eta}$ and $\Omega^{\prime}_{\iota,\eta, \rho}$ we can then finally define the normalizing vectors in terms of the plan as follows:

\begin{align}\label{eq:norm_vecs_plan}
    \mathcal{Z}^{\eta} = \frac{\sum_{[i,j]\in \Omega_{\iota,\eta, \rho}}P^{\iota}_{i,j}}{\sum_{[i,j]\in \mathcal{O}_{\iota,\eta, \rho}}P^{\iota}_{i,j}}, ~~ \mathcal{Z}^{\omega(\eta)} = \frac{\sum_{[i,j]\in \Omega^{\prime}_{\iota,\eta, \rho}}P^{\iota}_{i,j}}{\sum_{[i,j]\in \mathcal{O}^{\prime}_{\iota, \eta, \rho^{\prime}}}P^{\iota}_{i,j}}
\end{align}

We also present the derivation of the special case in which the intervened variables have \emph{no parents}. Specifically,  one can easily see that in this case:
$$\mathcal{O}_{\iota,\eta, \rho} = \left\{ [i,j]~ \mid ~x_j \in \dom{\mathbf{X}} \right\}~~ \text{for } \rho \in \dom{\operatorname{PA}_{\mathbf{B}}}$$
$$\mathcal{O}^{\prime}_{\iota, \eta, \rho^{\prime}} = \left\{ [i,j]~ \mid ~x^{\prime}_i \in \dom{\mathbf{X^{\prime}}} \right\}~~ \text{for } \rho^{\prime} \in \dom{\operatorname{PA}_{\mathbf{B^{\prime}}}}$$

This implies that $\Omega_{\iota,\eta, \rho} = \mathcal{C}_{\iota, \eta}$ and similarly $ \Omega^{\prime}_{\iota,\eta, \rho} = \mathcal{C}^{\prime}_{\iota, \eta}$. Therefore, since $\operatorname{PA}_{\mathbf{B}} = \operatorname{PA}_{\mathbf{B^{\prime}}} = \varnothing$ then $\cmpt{x_j}{\scmdo(\operatorname{PA}_{\mathbf{B}}=\rho)} ~\forall ~ x_j \in \dom{\mathbf{X}}$ and $\cmpt{x^{\prime}_i}{\scmdo(\operatorname{PA}_{\mathbf{B^{\prime}}}=\rho^{\prime})} ~\forall ~ x^{\prime}_i \in \dom{\mathbf{X^{\prime}}}$, which suggest that $\sum_{[i,j]\in \mathcal{O}_{\iota,\eta, \rho}}P^{\iota}_{i,j} = \sum_{[i,j]\in \mathcal{O}^{\prime}_{\iota, \eta, \rho^{\prime}}}P^{\iota}_{i,j} = 1$. Therefore, the normalizing vectors in \cref{eq:norm_vecs_plan} now become:
\begin{align}
    \mathcal{Z}^{\eta} =\sum_{[i,j]\in \mathcal{C}_{\iota, \eta}}P^{\iota}_{i,j}, ~~ \mathcal{Z}^{\omega(\eta)} =\sum_{[i,j]\in \mathcal{C}^{\prime}_{\iota, \eta}}P^{\iota}_{i,j}
\end{align}
The later relation conveys that in the case of \emph{no parents} for the intervened variable the normalizing vectors become constant vectors for each model and are not different for each of the $x_j \in \dom{\mathbf{X}}$ and $x^{\prime}_i \in \dom{\mathbf{X^{\prime}}}$ respectively, compared to the general case where we have a unique normalizing constant for each of these entries.
\section{Proof of Theorem 1}\label{sec:proof2}
Let $D = \dom{\mathbf{X}}$ and $D^\prime = \dom{\mathbf{X}^\prime}$. Let $N(k) = N$ (the number of plans depends on the chain $k$) for simplicity. Hence we have $N$ plans in a particular chain $\intervsetbase_k$. Also for simplicity of notation, let $P^{\iota_n} = P_n$, $C_{\omega} = C$.
We show the proof for a generic chain, and it holds for any chain in the set of chains.
To prove the theorem, we need to show:
\begin{enumerate}
    \item The set $\underbrace{\mathcal{S}_{D \times D^\prime} \times \dots \mathcal{S}_{D \times D^\prime}}_{N \text{times} } = \{ (P_1,\dots,P_N) ~ \mid ~ P_{n} \in \mathcal{S}_{D \times D^\prime}, $ for all $k=1,\dots,N \}$, where $\mathcal{S}_{D \times D^\prime}$ is the set of stochastic matrices of dimension $D \times D^\prime$, is a convex set.
    \item The constraints $\{ P_n \in \mathcal{U}(\pi_n), n=1,\dots,N \} $ are convex.
    \item The function $ (P_1,\dots,P_N) \to \sum_{n = 1}^{N} \kappa \cdot \Big<C,P_{n}\Big> + \bm{\lambda}^\top  \mathcal{D}\left(P_n;P_{n+1} \right) + \mu \cdot \mathcal{H}(P_{n})$ is jointly convex in $P_1,\dots,P_N$. Denote this function by $f^{\text{COTA}}(P_1,\dots,P_N)$.
\end{enumerate}
Point (1) is a consequence of the fact that each $P_n$ is in the set of stochastic matrices $\mathcal{S}_{D \times D^\prime}$, which is a convex set, and that the Cartesian product $\times$ preserves convexity. Point (2) follows since each $\mathcal{U}(\pi_n)$ is a convex set \citep{peyre2019computational}. To prove (3), we are going to use the following facts.
\begin{lemma*}
The domain of each $P_n$ is a convex set, and the inner product $P_n \to \Big<C,P_{n} \Big>$ is a convex function in $P_n$.  
\end{lemma*}
Proof: the domain of $P_n$ is the stochastic matrices as described in point (1) above; the inner product is a convex function in $P_n$ \citep{boyd2004convex} $\blacksquare$ \\ Further, $\bm{\lambda}^\top \mathcal{D}(P_n ; P_{n+1})$ is jointly convex in the pair $(P_n,P_{n+1})$, which the next lemma shows.

\begin{lemma*}
   The function $(P_n, P_{n+1}) \rightarrow \bm{\lambda}^\top \mathcal{D}(P_n ; P_{n+1})$ is jointly convex in the pair $(P_n, P_{n+1})$
\end{lemma*}
$$\mathcal{D}(P_n, P_{n+1}) = [\delta_{n,n+1}, \delta^\prime_{n,n+1}]^\top$$

Given that:
\begin{enumerate}
    \item \( P_{n}, P_{n+1} \) are stochastic matrices, i.e., they belong to the set \( \mathcal{S}_{D \times D^\prime} \).
    \item \( d \) is a Bregman divergence between probability vectors, which is jointly convex in both its inputs \citep{dhillon2008matrix}.
    \item \( \delta(P_{n}, P_{n+1}) = d(P_{n} \mathbf{1}, P_{n+1} \mathbf{1}) \) , \( \delta^\prime(P_{n}, P_{n+1}) = d(P_{n}^\top \mathbf{1}, P_{n+1}^\top \mathbf{1}) \) where $\mathbf{1}$ is a vector of ones of appropriate dimension. Note that $P_{n} \mathbf{1}$ and $P_{n}^\top \mathbf{1}$ are the marginals of $P_{n}$.
\end{enumerate}

\noindent We want to prove that $ \bm{\lambda}^\top \mathcal{D}(P_{n}, P_{n+1})$ is jointly convex in \( P_{n} \) and \( P_{n+1} \), i.e.,  for $P_{n}^\prime, P_{n+1}^\prime$ any stochastic matrices of the corresponding sizes and $\gamma \in [0,1]$, it holds (due also to linearity of matrix multiplication)
\begin{equation}  \label{eq:joint_convexity_pair}
   \bm{\lambda}^\top \mathcal{D}(\gamma P_{n} + (1-\gamma) P_{n}^\prime, \gamma P_{n+1} + (1-\gamma) P_{n+1}^\prime) \leq \gamma  \bm{\lambda}^\top \mathcal{D}(P_{n},P_{n+1}) + (1 - \gamma) \bm{\lambda}^\top \mathcal{D}(P_{n}^\prime ,P_{n+1}^\prime) . 
\end{equation}

It is now sufficient to prove that both $\delta(P_{n}, P_{n+1}), \delta^\prime(P_{n}, P_{n+1})$ are jointly convex in  \( P_{n} \) and \( P_{n+1} \), and convexity of $\bm{\lambda}^\top \mathcal{D}(\cdot,\cdot)$ follows. We prove the result for $\delta$ and an analogous proof holds for $\delta^\prime$. 

Expanding:
\[ \delta(\gamma P_{n} + (1-\gamma) P_{n}^\prime, \gamma P_{n+1} + (1-\gamma) P_{n+1}^\prime) = d((\gamma P_{n} + (1-\gamma) P_{n}^\prime) \mathbf{1}, (\gamma P_{n+1} + (1-\gamma) P_{n+1}^\prime) \mathbf{1}) , \] 
given that \( d \) is a Bregman divergence and is jointly convex, for any probability vectors \( \alpha, \alpha', \beta, \beta' \):
\[ d(\gamma \alpha + (1-\gamma) \alpha', \gamma \beta + (1-\gamma) \beta') \leq \gamma d(\alpha, \beta) + (1-\gamma) d(\alpha', \beta') \]

Setting \( \alpha = P_{n} \mathbf{1} \), \( \alpha' = P_{n}^\prime \mathbf{1} \), \( \beta = P_{n+1} \mathbf{1} \), and \( \beta' = P_{n+1}^\prime \mathbf{1} \), we get (simply by linearity of matrix-vector multiplication):
\[ d(\gamma P_{n} \mathbf{1} + (1-\gamma) P_{n}^\prime \mathbf{1}, \gamma P_{n+1} \mathbf{1} + (1-\gamma) P_{n+1}^\prime \mathbf{1}) \leq \gamma d(P_{n} \mathbf{1}, P_{n+1} \mathbf{1}) + (1-\gamma) d(P_{n}^\prime \mathbf{1}, P_{n+1}^\prime \mathbf{1}) . \]

This is equivalent to:
\[ \delta(\gamma P_{n} + (1-\gamma) P_{n}^\prime, \gamma P_{n+1} + (1-\gamma) P_{n+1}^\prime) \leq \gamma \delta(P_{n}, P_{n+1}) + (1-\gamma) \delta(P_{n}^\prime, P_{n+1}^\prime) . \]

Thus, if \( d \) is a Bregman divergence, making it therefore jointly convex in its two vector inputs, then \( \delta \) is jointly convex in its two matrix inputs \( P_{n} \) and \( P_{n+1} \). $\blacksquare$

\noindent Now, we resort to the definition of joint convexity of a function $f : \underbrace{\mathcal{S}_{D \times D^\prime} \times \dots \mathcal{S}_{D \times D^\prime}}_{N \text{times} } \to \mathbb{R}_{\geq 0}$ in the tuple of matrices $(P_1,\dots,P_N) \in \mathcal{S}_{D \times D^\prime} \times \dots \mathcal{S}_{D \times D^\prime}$. \\
\textbf{Definition 1} A function $f$ is jointly convex in $P_1,\dots,P_N$ iff, for $\gamma \in [0,1]$ and the domain of $(P_1,\dots,P_N)$ is a convex set, it holds
\begin{equation}
    f(\gamma P_1 + (1-\gamma)P_{1}^{\prime} , \dots, \gamma P_N + (1-\gamma)P_{N}^{\prime}) \leq \gamma f(P_1,\dots,P_N) + (1-\gamma) f(P_{1}^{\prime},\dots,P_{N}^{\prime}) .
\end{equation}
We will show that for $f^{\text{COTA}}(P_1,\dots,P_N)$, joint convexity follows from convexity of the inner product in $P_{n}$ and joint convexity in the pairs $(P_{n},P_{n+1})$ of the distance.\\

By taking the convex combinations $\gamma P_1 + (1-\gamma)P_{1}^{\prime} , \dots, \gamma P_N + (1-\gamma)P_{N}^{\prime}$ and plugging it in, we get: 
\begin{align}
    &f^{\text{COTA}}(\gamma P_1 + (1-\gamma)P_{1}^{\prime} , \dots, \gamma P_N + (1-\gamma)P_{N}^{\prime}) \\ &=\sum_{n = 1}^{N} \kappa \cdot \Big<C, \gamma P_{n} + (1-\gamma)P_{n}^{\prime} \Big> + \bm{\lambda}^\top  \mathcal{D}(\gamma P_{n} + (1-\gamma)P_{n}^{\prime} ~ ; ~ \gamma P_{n+1} + (1-\gamma)P_{n+1}^{\prime}) \\ & + \mu \cdot \mathcal{H}(\gamma P_{n} + (1-\gamma)P_{n}^{\prime}) \nonumber. 
\end{align}
By combining the fact that the inner product is convex in $P_{n}$ so $ \Big<C, \gamma P_{n} + (1-\gamma) P_{n}^{\prime} \Big> \leq \gamma  \Big<C, P_{n} \Big> + (1-\gamma)   \Big<C, P_{n}^{\prime} \Big> $, the inequality from \cref{eq:joint_convexity_pair}, and the (strict) convexity of $\mathcal{H}$, we have 
\begin{align}
      & f^{\text{COTA}}(\gamma P_1 + (1-\gamma)P_{1}^{\prime} , \dots, \gamma P_N + (1-\gamma)P_{N}^{\prime}) \\ 
      & \leq \sum_{n=1}^{N}  \kappa \cdot \left ( \gamma \Big<C, P_{n}  \Big> + (1-\gamma) \Big<C, P_{n}^{\prime}  \Big> \right ) + \bm{\lambda}^\top  \left ( \gamma  \mathcal{D}( P_{n} ; P_{n+1} ) + (1-\gamma) \mathcal{D}( P_{n}^{\prime} ; P_{n+1}^{\prime} ) \right )  \nonumber  \\ &+ \mu \cdot \left ( \gamma \mathcal{H}(P_{n}) + (1 - \gamma) \mathcal{H}(P_{n}^{\prime}) \right ) \nonumber \\
      &= \gamma \left ( \sum_{n=1}^{N} \kappa \Big<C, P_{n} \Big> + \lambda \mathcal{D}( P_{n} ; P_{n+1} ) + \mu \mathcal{H}(P_n)  \right ) + \left (1- \gamma \right ) \sum_{n=1}^{N}   \kappa \Big<C, P_{n}^{\prime} \Big> + \bm{\lambda}^\top \mathcal{D}( P_{n}^{\prime} ; P_{n+1}^{\prime} ) + \mu \mathcal{H}(P_{n}^{\prime})   \nonumber \\
      &= \gamma f^{\text{COTA}}(P_1,\dots,P_N) + (1-\gamma) f^{\text{COTA}}(P_{1}^{\prime},\dots,P_{N}^\prime) \nonumber
\end{align}
which is the definition of convexity in $P_1,\dots,P_N$.$\blacksquare$

\section{OT costs}\label{sec:costs}
\begin{figure*}
    \centering
	\begin{tikzpicture}[shorten >=1pt, auto, node distance=1cm, thick, scale=1.0, every node/.style={scale=1.0}, every left delimiter/.style={yshift=.9em,xshift=.5},
    every right delimiter/.style={yshift=.9em,xshift=.5}]
		
		\node (y1) at (-2.4,.7) {$\mathbf{x}^{\prime}_1 = 00$};
		\node (y2) at (-2.4,.25) {$\mathbf{x}^{\prime}_2 = 01$};
		\node (y3) at (-2.4,-.25) {$\mathbf{x}^{\prime}_3 = 10$};
		\node (y4) at (-2.4,-.7) {$\mathbf{x}^{\prime}_4 = 11$};
		
		\node[rotate=90] (x1) at (-1.1,1.9) {$\mathbf{x}_1 = 000$};
		\node[rotate=90] (x2) at (-.65,1.9) {$\mathbf{x}_2 = 001$};
		\node[rotate=90] (x3) at (-.2,1.9) {$\mathbf{x}_3 = 010$};
		\node[rotate=90] (x4) at (.25,1.9) {$\mathbf{x}_4 = 011$};
		\node[rotate=90] (x5) at (.7,1.9) {$\mathbf{x}_5= 110$};
		\node[rotate=90] (x6) at (1.15,1.9) {$\mathbf{x}_6 = 110$};

		\matrix at (0,0)  [{matrix of math nodes},left delimiter={[},right delimiter= {]}] (m)
		{
			0 & 0 & 1 & 1 & 2 & 2 & \\
			0 & 0 & 1 & 1 & 2 & 2 & \\
			2 & 2 & 2 & 2 & 2 & 2 &\\
			2 & 2 & 2 & 2 & 2 & 2 & \\		
		};  
		\draw[fill=red, color=green, fill opacity=0.2] (m-1-1.north west) -- (m-1-4.north east) -- (m-2-4.south east) -- (m-2-1.south west) -- (m-1-1.north west);
		\draw[fill=red, color=green, fill opacity=0.3] (m-1-1.north west) -- (m-1-2.north east) -- (m-2-2.south east) -- (m-2-1.south west) -- (m-1-1.north west);

		\node (I1) at (3.5,1.2) {$\iota_1: \scmdo(\emptyset) \overset{\omega}{\mapsto}  \scmdo(\emptyset)$};
		\node (I2) at (4.7,.6) {$\iota_2: \scmdo(\mathbf{X}[1] = 0) \overset{\omega}{\mapsto}  \scmdo(\mathbf{X}^\prime[1] = 0)$};
		\node (I3a) at (4.4,0) {$\iota_3: \scmdo(\mathbf{X}[1] = 0, \mathbf{X}[2] = 0) \overset{\omega}{\mapsto}$};
        \node (I3b) at (3.9,-.5) {$ \scmdo(\mathbf{X}^\prime[1] = 0)$};
		
	\end{tikzpicture}
 \captionsetup{skip=10pt}
    \caption{
    $\omega$-informed cost matrix computer with $|\intervsetbase|=3$. Notice, how all values $\mathbf{x}$ and $\mathbf{x}^{\prime}$ are compatible with $\iota_1$ and $\omega(\iota_1)$, while only $\mathbf{x}_1,\mathbf{x}_2$ are compatible with $\scmdo(\mathbf{X}[1] = 0, \mathbf{X}[2] = 0)$ and $\mathbf{x}^{\prime}_1$ is compatible with $\scmdo(\mathbf{X}^{\prime}[1] = 0)$
    }
    \label{fig:viz_omega}
\end{figure*}
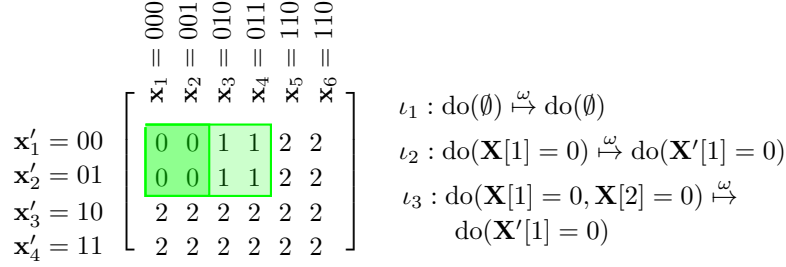

In this section we discuss the mechanics underlying the computations of the OT costs. First, we recall the $\omega$-cost function formula as it is introduced in the \cref{eq:omega_cost_func}. Given two samples $\mathbf{x} \in \dom{\mathbf{X}}$ and $\mathbf{x}^{\prime} \in \dom{\mathbf{X}^{\prime}}$, and an intervention set $\intervsetbase$ together with $\omega:\intervsetbase\to \intervsetabst$, we define the $\omega$-cost as the following function:
$$
    c_{\displaystyle \omega}(\mathbf{x},\mathbf{x}^\prime) = \cardinality{\intervsetbase} - \sum\nolimits_{\iota \in \intervsetbase} 
    \indicator \left[ \cmpt{\mathbf{x}}{\iota} \logicaland \cmpt{\mathbf{x}^\prime}{\omega(\iota)}\right]
$$
The function $c_{ \omega}$ discounts the cost of transporting samples $\mathbf{x}$ to $\mathbf{x^{\prime}}$ proportionally to the number of pairs $(\iota,\omega(\iota))$ w.r.t. which $\mathbf{x}$ and $\mathbf{x^{\prime}}$ are compatible. In \cref{fig:viz_omega} we visualise the construction of the cost matrix $C_{\omega}$ based on the $c_{\omega}(\cdot, \cdot)$ function of \cref{eq:omega_cost_func}.

As shown in \cref{sec:experiments}, for $c_{\omega}$ to be an effective cost function, the intervention set needs to be well-specified and informative regarding the domains of the two models. In such a case, a meaningful cost matrix can be derived and more elements on the induced transport plans will be influenced by causal knowledge. Our experiments confirmed that in scenarios where the intervention set was limited, conventional costs like Hamming $c_{\mathcal{H}}$ could return lower abstraction errors compared to $c_{\omega}$. The reason is that, by construction, $c_{\omega}$ will assign maximum values $(|\intervsetbase|)$ for samples for which it has no available information from the $\omega$ map. On the other hand, costs like $c_{\mathcal{H}}$ which spreads its values across the whole domain might be able to capture certain patterns more efficiently. We provide a visualization of the cost matrices for both $c_{\omega}$ and $c_{\mathcal{H}}$ functions in \cref{fig:omega_costs} and \cref{fig:ham_costs} respectively.

\begin{table}[t]
    \begin{minipage}[b]{0.49\linewidth}
        \resizebox{\columnwidth}{!}{%
        \includegraphics{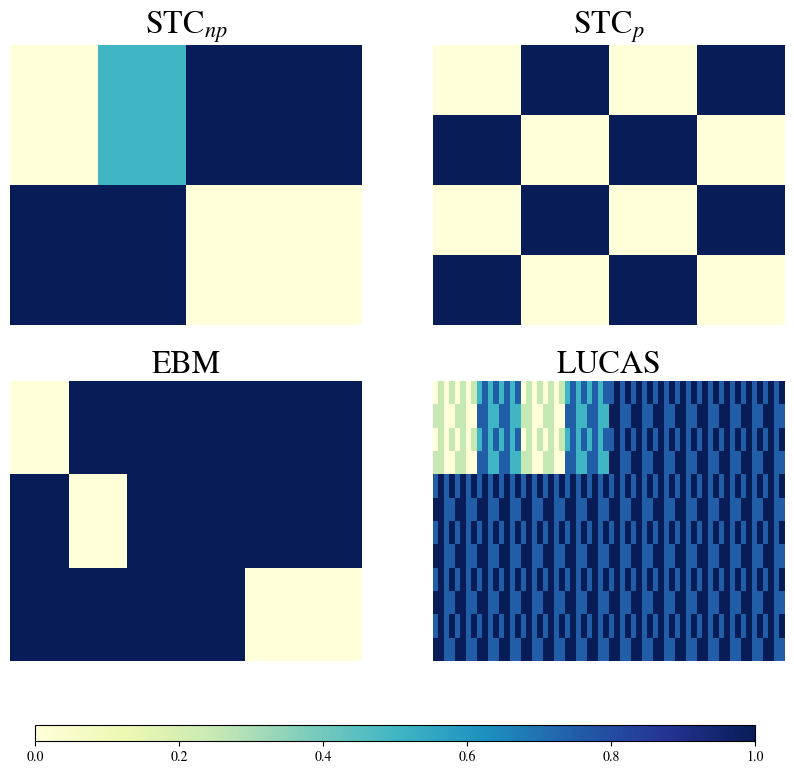}}
        \captionsetup{skip=10pt}
        \captionof{figure}{Omega Cost ($C_{\omega}$)}
        \label{fig:omega_costs}
    \end{minipage}\hfill%
    \begin{minipage}[b]{0.49\linewidth}
        \resizebox{\columnwidth}{!}{%
        \includegraphics{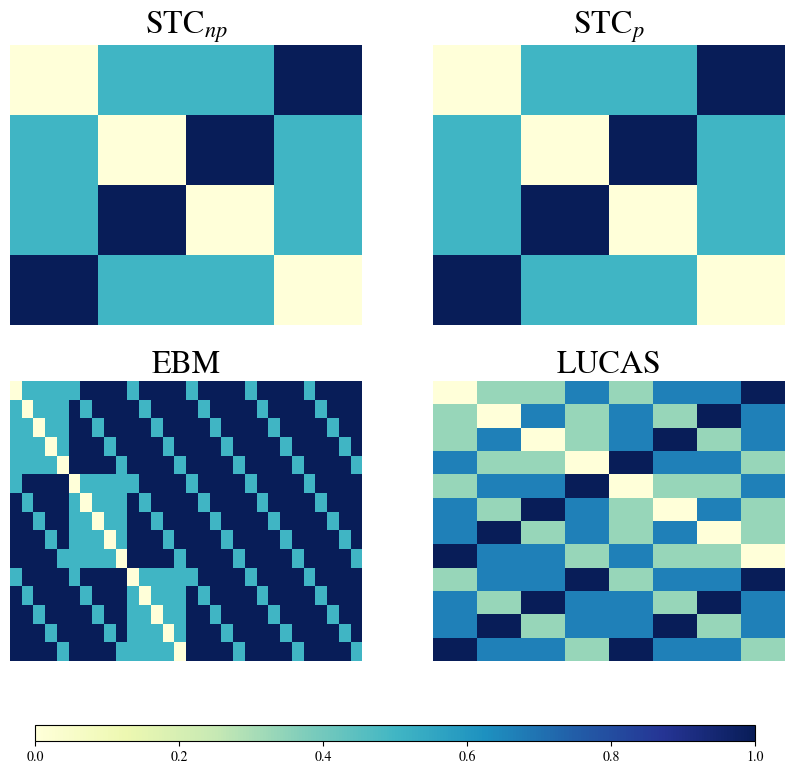}}
        \captionsetup{skip=10pt}
        \captionof{figure}{Hamming cost ($C_{\mathcal{H}}$)}
        \label{fig:ham_costs}
    \end{minipage}
\end{table}

\section{EBM Downstream Task}\label{sec:downstream_details}
In our \textbf{EBM} scenario we rely on data about battery coating released by the Laboratoire de Réactivité et Chimie des Solides (LRCS) \citep{cunha2020artificial} and by the Warwick Manufacturing Group (WMG) \citep{zennaro2023jointly}. These datasets contain observations about different variables affecting the coating process (\emph{Comma Gap, Mass Loading Position}), as well as observation about a key outcome variable related to the width of the coating (\emph{Mass Loading}). Both groups aim at inferring a machine learning model that would allow them to control the target variable via interventions on the other variables.

Closely following \cite{zennaro2023jointly}, we learn abstractions from the WMG model to the LRCS model in order to merge data collected by the two laboratories and learn a model from the aggregate dataset. We compare our approach \cota{} with the competing \ca{} learning method proposed in \citep{zennaro2023jointly}. The latter builds upon the $\mathbf{\alpha}$-abstraction framework established by \cite{rischel2020category}, which is briefly summarized in \cref{sec:alphaabstraction}.
To assess the usefulness of abstraction, we solve three extrapolation tasks meant to show how fitting a simple model to the aggregated dataset produced via abstraction guarantees better predictive results. We consider the following three downstream tasks:
\begin{itemize}
    \item In the first task, we train a regression model on all LRCS samples except for samples belonging to class $k$. We then test the regression model on LRCS samples belonging to class $k$. This constitutes the baseline model learned on data from a single laboratory.
    \item In the second task, we train a regression model on all LRCS samples except for samples belonging to class $k$ and all the abstracted WMG samples. We then test the regression model on LRCS samples belonging to class $k$. This constitutes a scenario in which we enrich one dataset (LRCS) with data from another laboratory (WMG); moreover, the enriching data provides information on the class $k$ which was not originally observerd in LRCS. 
    \item In the third task, we train a regression model on all LRCS and abstracted WMG samples except for samples belonging to class $k$. We then test the regression model on LRCS and WMG samples belonging to class $k$. This constitutes a scenario in which we enrich one dataset (LRCS) with data from another laboratory (WMG); however, the enriching data also lacks observations for the class $k$. 
\end{itemize}
Our solution outperforms the SOTA in terms of MSE, and confirms that using abstractions to aggregate data may be beneficial for downstream tasks, such as regression tasks. This is particularly true in settings where data are limited because of the cost and complexity of collecting samples, such as in the case of battery manufacturing \cite{Niri2022}.

\subsection{The $\mathbf{\alpha}$-abstraction framework}\label{sec:alphaabstraction}
This framework draws inspiration from category theory, and assumes two \scm{s} $\scmbase = \langle \mathbf{X},\mathbf{U},\structfuncset,\prob(\mathbf{U}) \rangle$, $\scmabst=\langle \mathbf{X^\prime},\mathbf{U^\prime},\structfuncset^\prime,\prob^\prime(\mathbf{U^\prime}) \rangle$ with finite sets of endogenous variables, where each variable defined on a finite and discrete domain.
\begin{definition*}[\cite{rischel2020category}]
    Given two \scm{s} $\scmbase$ and $\scmabst$, an abstraction $\mathbf{\alpha}$ is a tuple $\left<R, a, \mathbf{\alpha} \right>$ where:
    \begin{itemize}[itemsep=0.01pt]
        \item $R\subseteq \mathbf{X}$ is a subset of relevant variables in the model $\scmbase$.
        \item $a:R\to \mathbf{X^\prime}$ is a surjective map between variables, from nodes in $\scmbase$ to nodes in $\scmabst$.
        \item $\mathbf{\alpha}$ is a collection of surjective maps $\alpha_{X^\prime}: \dom{a^{-1}(X^\prime)} \to \dom{X^\prime}$ where $X \subseteq \mathbf{X}$ and $X^\prime\subseteq \mathbf{X^\prime}$.
    \end{itemize}
\end{definition*}
An $\mathbf{\alpha}$-abstraction establishes an asymmetric relation from a base model $\scmbase$ to an abstracted model $\scmabst$. This definition encodes a mapping on two layers: on a structural or graphical level between the nodes of the \dirag{s} via $a$, and on a distributional level via the maps
$\alpha_{X^\prime}$. 

\cite{rischel2020category} introduces a notion of interventional consistency between base and abstracted model, whereby interventional distributions produced in the base and abstracted model are related via the abstraction $\mathbf{\alpha}$; furthermore, a notion of abstraction error, analogous to the one used in this paper, is also proposed. It is then immediate to relate the $\tau$-abstraction and $\alpha$-abstraction as they both imply comparing distributions generated by base and abstracted model through an abstraction map.

\section{Additional Experiments and Analysis}\label{sec:complete_results}
Here we report the complete experimental results of our simulations. These results corroborate our understanding of the effectiveness of \cota{} compared to the baseline methods. Regarding the parameter $\bm{\lambda}^\top$, in \cref{sec:single_lambda}  we demonstrate the complete results from the the equal weight case where $\lambda =\lambda^{\prime}$ and in \cref{sec:multi_lambda} from the more general case of $\lambda \neq \lambda^{\prime}$. Notice, that throughout the results, how in \textbf{STC}$_{\text{np}}$ and \textbf{LUCAS} $\omega$-cost $c_{\omega}$ returns lower abstraction errors compared to Hamming $c_{\mathcal{H}}$, whereas in \textbf{STC}$_{\text{p}}$ and \textbf{EBM} the opposite holds true due to the smaller and less diverse intervention sets, as discussed in the main text and in \cref{sec:costs}. 

\subsection{Equal weights $(\lambda = \lambda^{\prime})$.}\label{sec:single_lambda}
We present the evaluation results in the case where the parameter $\bm{\lambda}^\top$ governing the causal constraint term in the \cota{} optimization objective of \cref{eq:objective_2d} is a constant vector i.e. $\bm{\lambda}^\top \cdot\mathcal{D}(P^{\iota}, P^{\eta}) = [\lambda, \lambda]\cdot \begin{bmatrix}
\delta_{\iota, \eta}  \\
\delta^{\prime}_{\iota, \eta}
\end{bmatrix}= \lambda\cdot(\delta_{\iota, \eta}+\delta_{\iota, \eta}^{\prime})$. The complete evaluation for the \textbf{STC} examples is detailed in \autoref{tab:np_gc_all} and \autoref{tab:p_gc_all}, while results for the \textbf{LUCAS} and the \textbf{EBM} examples are presented in \autoref{tab:lucas_gc_all} and \autoref{tab:ebm_gc_all} respectively.

\subsection{Different weights $(\lambda \neq \lambda^{\prime})$.}\label{sec:multi_lambda}
We present the evaluation results in the case where the parameter $\bm{\lambda}^\top$ governing the causal constraint term in the \cota{} optimization objective of \cref{eq:objective_2d} is a non-constant vector i.e. $\bm{\lambda}^\top \cdot \mathcal{D}(P^{\iota}, P^{\eta}) = \left[\lambda ~~ \lambda^{\prime}\right] \cdot \begin{bmatrix}
           \delta_{\iota, \eta}  \\
           \delta^{\prime}_{\iota, \eta}
         \end{bmatrix} =\lambda\cdot\delta_{\iota, \eta}+\lambda^{\prime}\cdot\delta_{\iota, \eta}^{\prime}$. The complete evaluation for the \textbf{STC} examples is detailed in \autoref{tab:np_gc_all_multi} and \autoref{tab:p_gc_all_multi}, while results for the \textbf{LUCAS} and the \textbf{EBM} examples are presented in \autoref{tab:lucas_gc_all_multi} and \autoref{tab:ebm_gc_all_multi}, respectively. Finally, \autoref{tab:downstream_multi} demonstrates the MSE of \cota{} and a SOTA \ca{} framework on a regression task for \textbf{EBM}.

\section{\texttt{Approximate} \cota{}}\label{sec:approx_cota}
We can simplify the optimization problem and halve the number of constraints by assuming that the elements of each marginal of the plan inherit the marginal’s normalising factor (Occam’s razor). This way, we turn $\delta_{\iota, \eta}$ and  $\delta^{\prime}_{\iota, \eta}$ into the element-wise (between the plans) distances $d\left( \left( \frac{1}{\mathcal{Z}^{\eta}_{j}} P^{\iota}_{ij} \right)_{ij}, \left( P^{\eta}_{ij} \right)_{ij}\right)$ and $d\left( \left(\frac{1}{\mathcal{Z}^{\omega(\eta)}_{i}}P^{\iota}_{i j}\right)_{ij}, \left(P^{\eta}_{i j}\right)_{ij}\right)$, respectively. Now every pair of elements $P^{\iota}_{ij},P^{\eta}_{ij}$ is related simultaneously through $\mathcal{Z}^{\eta}_j$ and $\mathcal{Z}^{\omega(\eta)}_i$. Then given $d:\mathbb{R}^{D\times D^{\prime}}\times \mathbb{R}^{D\times D^{\prime}}\to  \mathbb{R}_{\geq 0}$, we re-express our constraints in a matrix form as:
\begin{equation}\label{eq:constraint_tensor_plans}
    \mathcal{D}(P^{\iota}, P^{\eta}) = d\left( \frac{1}{\varphi(\mathcal{Z}^{\eta}_j, \mathcal{Z}^{\omega(\eta)}_i)} P^{\iota}, P^{\eta}\right) \quad \text{ if }\cmpt{x_j}{\eta} \text{ and } \cmpt{x_i^{\prime}}{\omega(\eta)}
\end{equation}
where $\varphi(\cdot, \cdot): \mathbb{R}^2 \rightarrow \mathbb{R}$ is any aggregating function that preserves the correct support of the plans. In our experiments we work with $\varphi(\mathcal{Z}^{\eta}_j, \mathcal{Z}^{\omega(\eta)}_i) = \min(\mathcal{Z}^{\eta}_j, \mathcal{Z}^{\omega(\eta)}_i)$. We provide the complete abstraction error evaluation results as before alongside the downstream evaluation for the \texttt{Approximate} \cota{} in \cref{sec:approx_cota_results}.

\subsection{\texttt{Approximate} \cota{} Results}\label{sec:approx_cota_results}
We report the complete experimental results for \texttt{Approximate} \cota{}. Once again, we confirm our understanding of the effectiveness of \cota{} compared to the baseline methods. The complete evaluation for the \textbf{STC} examples is detailed in \autoref{tab:STC_NP_results_all} and \autoref{tab:STC_P_results_all}, while results for the \textbf{LUCAS} and the \textbf{EBM} examples are presented in \autoref{tab:LUCAS_results_all} and \autoref{tab:EBM_results_all}, respectively. Furthermore, in \textbf{STC}$_{\text{np}}$ and \textbf{LUCAS} $\omega$-cost $c_{\omega}$ returns lower abstraction errors compared to Hamming $c_{\mathcal{H}}$, whereas in \textbf{STC}$_{\text{p}}$ and \textbf{EBM} the opposite holds true due to the smaller and less diverse intervention sets, as discussed in the main text and in \cref{sec:costs}. In addition, in \cref{fig:fro_ternary} and \cref{fig:jsd_ternary} we provide the equivalent simplex plots for \textbf{STC}$_{\text{np}}$, which illustrate the influence of the parameter $\lambda$ in the optimization problem. Once again, these plots reinforce the idea that optimal solutions are achieved when $\lambda$ is greater than 0. Finally, in \autoref{tab:downstream_approx_cota}, we demonstrate the results regarding the MSE for the downstream task on the \textbf{EBM} dataset as before.

\vspace{5mm}
\begin{table}[H]
\begin{minipage}[b]{0.49\linewidth}
\resizebox{\columnwidth}{!}{%
\begin{tabular}{ccccc}
    \toprule
    Method & $\mathcal{D}$  & $\mathcal{C}$ & $e_{\jsd}(\tau)$   & $e_{\wass}(\tau)$\\
    \midrule
     \cota{($\widehat{P}$)} & \fro  & $c_{\omega}$ & $\bm{0.010 \pm 0.005}$&$\bm{0.011 \pm 0.003}$\\
     &  & $c_{\mathcal{H}}$ & $0.087 \pm 0.007$&$0.025 \pm 0.001$\\
     \\
     & \jsd  & $c_{\omega}$ & $0.012 \pm 0.006$&$0.012 \pm 0.003$\\
     &   & $c_{\mathcal{H}}$ & $0.087 \pm 0.006$ &$0.025 \pm 0.001$\\
     \\
    \midrule
     \cota{($\widehat{\tau}$)} & \fro  & $c_{\omega}$ & $0.013 \pm 0.021$ &$0.171 \pm 0.001$\\
     &  & $c_{\mathcal{H}}$ & $0.169 \pm 0.005$ &$0.178 \pm 0.001$\\
     \\
     & \jsd  & $c_{\omega}$ & $0.014 \pm 0.021$&$0.171 \pm 0.001$\\
     &   & $c_{\mathcal{H}}$ & $0.169 \pm 0.004$ &$0.178 \pm 0.001$\\
     \\
    \midrule
    \pairOT & --   & $c_{\omega}$    & $0.013 \pm 0.002$ &$0.011 \pm 0.002$\\
    & -- &    $c_{\mathcal{H}}$  & $0.093 \pm 0.004$ &$0.039 \pm 0.002$\\
    \midrule
    \mapOT{} & -- &  $c_{\omega}$    & $0.023 \pm 0.022$ &$0.147 \pm 0.001$\\
           & -- &  $c_{\mathcal{H}}$  & $0.169 \pm 0.022$ &$0.156 \pm 0.001$\\
    \midrule
    \baryOT{} & -- &  $c_{\omega}$    & $0.233 \pm 0.142$ &$0.067 \pm 0.042$\\
              & -- &  $c_{\mathcal{H}}$  & $0.323 \pm 0.074$ &$0.095 \pm 0.039$\\
    \bottomrule
        \end{tabular}}
        \captionsetup{skip=10pt}
\caption{{Complete abstraction error evaluation for the \textbf{STC}$_{\text{np}}$ example of the $\lambda = \lambda^{\prime}$ case. The $\cota{(\widehat{P})}-\fro - c_{\omega}$ returns the lower abstraction error for both metrics. The $\omega$-cost $c_{\omega}$ outperforms $c_{\mathcal{H}}$.}}
\label{tab:np_gc_all}
\end{minipage}\hfill%
\begin{minipage}[b]{0.49\linewidth}
\resizebox{\columnwidth}{!}{%

\begin{tabular}{ccccc}
    \toprule
    Method & $\mathcal{D}$  & $\mathcal{C}$ & $e_{\jsd}(\tau)$   & $e_{\wass}(\tau)$\\
    \midrule
     \cota{($\widehat{P}$)} & \fro  & $c_{\omega}$ & $\bm{0.010 \pm 0.006}$&$\bm{0.010 \pm 0.003}$\\
     &  & $c_{\mathcal{H}}$ & $0.087 \pm 0.006$&$0.025 \pm 0.001$\\
     \\
     & \jsd  & $c_{\omega}$ & $0.011 \pm 0.006$&$0.012 \pm 0.003$\\
     &   & $c_{\mathcal{H}}$ & $0.087 \pm 0.006$ &$0.025 \pm 0.001$\\
     \\
    \midrule
     \cota{($\widehat{\tau}$)} & \fro  & $c_{\omega}$ & $0.012 \pm 0.020$ &$0.171 \pm 0.001$\\
     &  & $c_{\mathcal{H}}$ & $0.169 \pm 0.004$ &$0.178 \pm 0.001$\\
     \\
     & \jsd  & $c_{\omega}$ & $0.013 \pm 0.019$&$0.171 \pm 0.001$\\
     &   & $c_{\mathcal{H}}$ & $0.169 \pm 0.004$ &$0.178 \pm 0.001$\\
     \\
    \midrule
    \pairOT & --   & $c_{\omega}$    & $0.013 \pm 0.002$ &$0.011 \pm 0.002$\\
    & -- &    $c_{\mathcal{H}}$  & $0.093 \pm 0.004$ &$0.039 \pm 0.002$\\
    \midrule
    \mapOT{} & -- &  $c_{\omega}$    & $0.023 \pm 0.022$ &$0.147 \pm 0.001$\\
           & -- &  $c_{\mathcal{H}}$  & $0.169 \pm 0.022$ &$0.156 \pm 0.001$\\
    \midrule
    \baryOT{} & -- &  $c_{\omega}$    & $0.233 \pm 0.142$ &$0.067 \pm 0.042$\\
              & -- &  $c_{\mathcal{H}}$  & $0.323 \pm 0.074$ &$0.095 \pm 0.039$\\
    \bottomrule
        \end{tabular}}
        \captionsetup{skip=10pt}
\caption{{Complete abstraction error evaluation for the \textbf{STC}$_{\text{np}}$ example of the $\lambda \neq \lambda^{\prime}$ case. The $\cota{(\widehat{P})}-\fro - c_{\omega}$ returns the lower abstraction error for both metrics. The $\omega$-cost $c_{\omega}$ outperforms $c_{\mathcal{H}}$.}}
\label{tab:np_gc_all_multi}
\end{minipage}
\end{table}

\newpage
\begin{table}
\begin{minipage}[b]{0.49\linewidth}
\resizebox{\columnwidth}{!}{%
\begin{tabular}{ccccc}
    \toprule
    Method & $\mathcal{D}$  & $\mathcal{C}$ & $e_{\jsd}(\tau)$   & $e_{\wass}(\tau)$\\
    \midrule
     \cota{($\widehat{P}$)} & \fro  & $c_{\omega}$ & $0.278\pm 0.015$&$\bm{0.048\pm 0.007}$\\
     &  & $c_{\mathcal{H}}$ & $0.241\pm 0.003$ &$\bm{0.048\pm 0.007}$\\
     \\
     & \jsd  & $c_{\omega}$ & $0.258\pm 0.027$&$0.054\pm 0.003$\\
     &   & $c_{\mathcal{H}}$ & $0.242\pm 0.001$ &$0.054\pm 0.003$\\
     \\
    \midrule
     \cota{($\widehat{\tau}$)} & \fro  & $c_{\omega}$ & $0.249\pm 0.005$ &$0.135 \pm 0.001$\\
     &  & $c_{\mathcal{H}}$ & $\bm{0.229\pm 0.003}$&$0.129 \pm 0.001$ \\
     \\
     & \jsd  & $c_{\omega}$ & $0.241\pm 0.008$&$0.135 \pm 0.001$\\
     &   & $c_{\mathcal{H}}$ & $\bm{0.229\pm 0.004}$ &$0.129\pm 0.001 $\\
     \\
    \midrule
    \pairOT & --   & $\omega$    & $0.279 \pm 0.014$ &$0.091 \pm 0.005$\\
    & -- &    $\mathcal{H}$  & $0.242 \pm 0.002$ &$0.067 \pm 0.001$\\
    \midrule
    \mapOT{} & -- &  $\omega$    & $0.250 \pm 0.005$ &$0.140 \pm 0.001$\\
           & -- &  $\mathcal{H}$  & $0.229 \pm 0.004$ &$0.129 \pm 0.001$\\
    \midrule
    \baryOT{} & -- &  $\omega$    & $0.318 \pm 0.205$ &$0.104 \pm 0.061$\\
              & -- &  $\mathcal{H}$  & $0.272 \pm 0.212$ &$0.075 \pm 0.058$\\
    \bottomrule
        \end{tabular}}
        \captionsetup{skip=10pt}
\caption{{
Complete abstraction error evaluation for the \textbf{STC}$_{\text{p}}$ example of the $\lambda = \lambda^{\prime}$ case. The $c_{\mathcal{H}}$ settings of $\cota{(\widehat{\tau})}$ formulation return the lower abstraction error for the \jsd{} metric and $\cota{(\widehat{P})}-\fro$ for the \wass{} metric. Table illustrates that $c_{\mathcal{H}}$ outperforms $c_{\omega}$ due to under-specification of the intervention set.}}
\label{tab:p_gc_all}
\end{minipage}\hfill%
\begin{minipage}[b]{0.49\linewidth}
\resizebox{\columnwidth}{!}{%
\begin{tabular}{ccccc}
        \toprule
    Method & $\mathcal{D}$  & $\mathcal{C}$ & $e_{\jsd}(\tau)$   & $e_{\wass}(\tau)$\\
    \midrule
     \cota{($\widehat{P}$)} & \fro  & $c_{\omega}$ & $0.277\pm 0.018$&$\bm{0.046\pm 0.007}$\\
     &  & $c_{\mathcal{H}}$ & $0.241\pm 0.002$ &$\bm{0.046\pm 0.007}$\\
     \\
     & \jsd  & $c_{\omega}$ & $0.273\pm 0.009$&$0.047\pm 0.006$\\
     &   & $c_{\mathcal{H}}$ & $0.241\pm 0.003$ &$0.047\pm 0.006$\\
     \\
    \midrule
     \cota{($\widehat{\tau}$)} & \fro  & $c_{\omega}$ & $0.249\pm 0.003$ &$0.135 \pm 0.004$\\
     &  & $c_{\mathcal{H}}$ & $\bm{0.228\pm 0.004}$&$0.129 \pm 0.001$ \\
     \\
     & \jsd  & $c_{\omega}$ & $0.247\pm 0.004$&$0.135 \pm 0.004$\\
     &   & $c_{\mathcal{H}}$ & $\bm{0.228\pm 0.003}$ &$0.129\pm 0.001 $\\
     \\
    \midrule
    \pairOT & --   & $\omega$    & $0.279 \pm 0.014$ &$0.091 \pm 0.005$\\
    & -- &    $\mathcal{H}$  & $0.242 \pm 0.002$ &$0.067 \pm 0.001$\\
    \midrule
    \mapOT{} & -- &  $\omega$    & $0.250 \pm 0.005$ &$0.140 \pm 0.001$\\
           & -- &  $\mathcal{H}$  & $0.229 \pm 0.004$ &$0.129 \pm 0.001$\\
    \midrule
    \baryOT{} & -- &  $\omega$    & $0.318 \pm 0.205$ &$0.104 \pm 0.061$\\
              & -- &  $\mathcal{H}$  & $0.272 \pm 0.212$ &$0.075 \pm 0.058$\\
    \bottomrule
        \end{tabular}}
        \captionsetup{skip=10pt}
\caption{{Complete abstraction error evaluation for the \textbf{STC}$_{\text{p}}$ example of the $\lambda \neq \lambda^{\prime}$ case. The $c_{\mathcal{H}}$ settings of $\cota{(\widehat{\tau})}$ formulation return the lower abstraction error for the \jsd{} metric and $\cota{(\widehat{P})}-\fro$ for the \wass{} metric. Table illustrates that $c_{\mathcal{H}}$ outperforms $c_{\omega}$ due to under-specification of the intervention set.}}
\label{tab:p_gc_all_multi}
\end{minipage}
\end{table}

\begin{table}
\begin{minipage}[b]{0.49\linewidth}
\resizebox{\columnwidth}{!}{%
\begin{tabular}{ccccc}
    \toprule
    Method & $\mathcal{D}$  & $\mathcal{C}$ & $e_{\jsd}(\tau)$   & $e_{\wass}(\tau)$\\
    \midrule
     \cota{($\widehat{P}$)} & \fro  & $c_{\omega}$ & $0.287 \pm 0.014$&$\bm{0.044 \pm 0.001}$\\
     &  & $c_{\mathcal{H}}$ & $0.287 \pm 0.014$&$0.047 \pm 0.001$\\
     \\
     & \jsd  & $c_{\omega}$ & $0.286 \pm 0.014$&$0.048 \pm 0.001$\\
     &   & $c_{\mathcal{H}}$ & $0.287 \pm 0.014$ &$0.048 \pm 0.001$\\
     \\
    \midrule
     \cota{($\widehat{\tau}$)} & \fro  & $c_{\omega}$ & $\bm{0.263 \pm 0.005}$ &$0.061 \pm 0.001$\\
     &  & $c_{\mathcal{H}}$ & $\bm{0.263 \pm 0.006}$ &$0.061 \pm 0.001$\\
     \\
     & \jsd  & $c_{\omega}$ & $\bm{0.263 \pm 0.005}$&$0.062 \pm 0.001$\\
     &   & $c_{\mathcal{H}}$ & $\bm{0.263 \pm 0.006}$ &$0.062 \pm 0.001$\\
     \\
    \midrule
    \pairOT & --   & $c_{\omega}$    & $0.306 \pm 0.009$ &$0.045 \pm 0.001$\\
    & -- &    $c_{\mathcal{H}}$  & $0.387 \pm 0.002$ &$0.047 \pm 0.001$\\
    \midrule
    \mapOT{} & -- &  $c_{\omega}$    & $0.294 \pm 0.008$ &$0.054 \pm 0.001$\\
           & -- &  $c_{\mathcal{H}}$  & $0.350 \pm 0.005$ &$0.054 \pm 0.001$\\
    \midrule
    \baryOT{} & -- &  $c_{\omega}$    & $0.294 \pm 0.047$ &$0.044 \pm 0.003$\\
              & -- &  $c_{\mathcal{H}}$  & $0.414 \pm 0.040$ &$0.046 \pm 0.010$\\
    \bottomrule
        \end{tabular}}
        \captionsetup{skip=10pt}
\caption{{Complete abstraction error evaluation for the \textbf{LUCAS} example of the $\lambda = \lambda^{\prime}$ case. All the settings of $\cota{(\widehat{\tau})}$ formulation return the lowest abstraction error for the \jsd{} metric and the $\cota{(\widehat{P})}-\fro - c_{\omega}$ setting for the \wass{} metric. The $\omega$-cost $c_{\omega}$ outperforms $c_{\mathcal{H}}$.}}
\label{tab:lucas_gc_all}
\end{minipage}\hfill%
\begin{minipage}[b]{0.49\linewidth}
\resizebox{\columnwidth}{!}{%
\begin{tabular}{ccccc}
        \toprule
    Method & $\mathcal{D}$  & $\mathcal{C}$ & $e_{\jsd}(\tau)$   & $e_{\wass}(\tau)$\\
    \midrule
     \cota{($\widehat{P}$)} & \fro  & $c_{\omega}$ & $0.287 \pm 0.014$&$ \bm{0.044\pm 0.001}$\\
     &  & $c_{\mathcal{H}}$ & $0.287 \pm 0.014$&$0.047 \pm 0.001$\\
     \\
     & \jsd  & $c_{\omega}$ & $0.287 \pm 0.014$&$0.045 \pm 0.001$\\
     &   & $c_{\mathcal{H}}$ & $0.287 \pm 0.014$ &$0.047 \pm 0.001$\\
     \\
    \midrule
     \cota{($\widehat{\tau}$)} & \fro  & $c_{\omega}$ & $\bm{0.263 \pm 0.005}$ &$0.060 \pm 0.001$\\
     &  & $c_{\mathcal{H}}$ & $\bm{0.263 \pm 0.006}$ &$0.060 \pm 0.002$\\
     \\
     & \jsd  & $c_{\omega}$ & $\bm{0.263 \pm 0.005}$&$0.061 \pm 0.001$\\
     &   & $c_{\mathcal{H}}$ & $\bm{0.263 \pm 0.006}$ &$0.061 \pm 0.001$\\
     \\
    \midrule
    \pairOT & --   & $c_{\omega}$    & $0.306 \pm 0.009$ &$0.045 \pm 0.001$\\
    & -- &    $c_{\mathcal{H}}$  & $0.387 \pm 0.002$ &$0.047 \pm 0.001$\\
    \midrule
    \mapOT{} & -- &  $c_{\omega}$    & $0.294 \pm 0.008$ &$0.054 \pm 0.001$\\
           & -- &  $c_{\mathcal{H}}$  & $0.350 \pm 0.005$ &$0.054 \pm 0.001$\\
    \midrule
    \baryOT{} & -- &  $c_{\omega}$    & $0.294 \pm 0.047$ &$0.044 \pm 0.003$\\
              & -- &  $c_{\mathcal{H}}$  & $0.414 \pm 0.040$ &$0.046 \pm 0.010$\\
    \bottomrule
        \end{tabular}}
        \captionsetup{skip=10pt}
\caption{{Complete abstraction error evaluation for the \textbf{LUCAS} example of the $\lambda \neq \lambda^{\prime}$ case. All the settings of $\cota{(\widehat{\tau})}$ formulation return the lowest abstraction error for the \jsd{} metric and the $\cota{(\widehat{P})}-\fro - c_{\omega}$ setting for the \wass{} metric. The $\omega$-cost $c_{\omega}$ outperforms $c_{\mathcal{H}}$.}}
\label{tab:lucas_gc_all_multi}
\end{minipage}
\end{table}

\begin{table}[b]
\begin{minipage}[b]{0.49\linewidth}
\resizebox{\columnwidth}{!}{%
\scalebox{.7}{
\begin{tabular}{ccccc}
    \toprule
    Method & $\mathcal{D}$  & $\mathcal{C}$ & $e_{\jsd}(\tau)$   & $e_{\wass}(\tau)$\\
    \midrule
     \cota{($\widehat{P}$)} & \fro  & $c_{\omega}$ &  $ 0.379$&$\bm{0.006}$\\
     &  & $c_{\mathcal{H}}$ & $0.263$ &$\bm{0.006}$\\
     \\
     & \jsd  & $c_{\omega}$ & $0.379$&$\bm{0.006}$\\
     &   & $c_{\mathcal{H}}$ & $0.263$ &$\bm{0.006}$\\
     \\
    \midrule
     \cota{($\widehat{\tau}$)} & \fro  & $c_{\omega}$ & $0.379$ &$0.053 $\\
     &  & $c_{\mathcal{H}}$ & $\bm{0.220}$&$0.053 $ \\
     \\
     & \jsd  & $c_{\omega}$ & $0.399$&$0.053 $\\
     &   & $c_{\mathcal{H}}$ & $0.263$ &$0.053 $\\
     \\
    \midrule
    \pairOT & --   & $c_{\omega}$    & $0.430$ &$0.027 $\\
    & -- &    $c_{\mathcal{H}}$  & $0.263$ &$0.027 $\\
    \midrule
    \mapOT{} & -- &  $c_{\omega}$    & $0.408$ &$0.060 $\\
           & -- &  $c_{\mathcal{H}}$  & $0.228$ &$0.053 $\\
    \midrule     
    \baryOT{} & -- &  $c_{\omega}$    & $0.530 $ &$0.019 $\\
              & -- &  $c_{\mathcal{H}}$  & $0.335$ &$0.023 $\\
    \bottomrule
        \end{tabular}}}
        \captionsetup{skip=10pt}
\caption{{Complete abstraction error evaluation for the \textbf{EBM} example of the $\lambda = \lambda^{\prime}$ case. The $\cota{(\widehat{\tau})}-\fro - c_{\mathcal{H}}$ formulation returns the lower abstraction error for the \jsd{} metric and all the settings of $\cota{(\widehat{P})}$ for the \wass{} metric. Table illustrates that $c_{\mathcal{H}}$ outperforms $c_{\omega}$ due to under-specification of the intervention set.}}
\label{tab:ebm_gc_all}
\end{minipage}\hfill%
\begin{minipage}[b]{0.49\linewidth}
\resizebox{\columnwidth}{!}{%
\scalebox{.7}{
\begin{tabular}{ccccc}
    \toprule
    Method & $\mathcal{D}$  & $\mathcal{C}$ & $e_{\jsd}(\tau)$   & $e_{\wass}(\tau)$\\
    \midrule
     \cota{($\widehat{P}$)} & \fro  & $c_{\omega}$ & $0.311$&$\bm{0.006}$\\
     &  & $c_{\mathcal{H}}$ & $0.263$ &$\bm{0.006}$\\
     \\
     & \jsd  & $c_{\omega}$ & $0.311$&$\bm{0.006}$\\
     &   & $c_{\mathcal{H}}$ & $0.263$ &$\bm{0.006}$\\
     \\
    \midrule
     \cota{($\widehat{\tau}$)} & \fro  & $c_{\omega}$ & $0.369$ &$0.053 $\\
     &  & $c_{\mathcal{H}}$ & $\bm{0.219}$&$0.053 $ \\
     \\
     & \jsd  & $c_{\omega}$ & $0.389$&$0.053 $\\
     &   & $c_{\mathcal{H}}$ & $0.220$ &$0.053 $\\
     \\
    \midrule
    \pairOT & --   & $c_{\omega}$    & $0.430$ &$0.027 $\\
    & -- &    $c_{\mathcal{H}}$  & $0.263$ &$0.027 $\\
    \midrule
    \mapOT{} & -- &  $c_{\omega}$    & $0.408$ &$0.060 $\\
           & -- &  $c_{\mathcal{H}}$  & $0.228$ &$0.053 $\\
    \midrule     
    \baryOT{} & -- &  $c_{\omega}$    & $0.530 $ &$0.019 $\\
              & -- &  $c_{\mathcal{H}}$  & $0.335$ &$0.023 $\\
    \bottomrule
        \end{tabular}}}
        \captionsetup{skip=10pt}
\caption{{Complete abstraction error evaluation for the \textbf{EBM} example of the $\lambda \neq \lambda^{\prime}$ case.  The $\cota{(\widehat{\tau})}-\fro - c_{\mathcal{H}}$ formulation returns the lower abstraction error for the \jsd{} metric and all the settings of $\cota{(\widehat{P})}$ for the \wass{} metric. Table illustrates that $c_{\mathcal{H}}$ outperforms $c_{\omega}$ due to under-specification of the intervention set.}}
\label{tab:ebm_gc_all_multi}
\end{minipage}
\end{table}

\begin{table}
    \centering
    \scalebox{1}{
  \begin{tabular}{cccc}
    \toprule
      Training set &   Test set &  \cite{zennaro2023jointly} &  \cota{}\\
    \midrule \\
    LRCS[$CG\neq k$] &  LRCS[$CG = k$] &  $1.86 \pm 1.75$ &  \bm{$1.40 \pm 1.39$} \\ 
    \midrule
   LRCS[$CG\neq k$] &  LRCS[$CG = k$] & $0.22 \pm 0.26$ & \bm{$0.20\pm0.02$ }\\
    +WMG & & \\  
    \midrule
     LRCS[$CG\neq k$] &  LRCS[$CG = k$] &  $1.22 \pm 0.95$ & \bm{ $0.48\pm0.23$ }\\ 
      +WMG[$CG\neq k$] &  WMG[$CG = k$] &  \\
    \bottomrule
  \end{tabular}
  }
      \captionsetup{skip=10pt}
\caption{MSE of \cota{} with $\lambda \neq \lambda^{\prime}$ and a SOTA \ca{} framework on a regression task for \textbf{EBM}. Augmenting
data via the learned abstraction reduces the average error in all different settings compared to the SOTA. We used $\cota{(\widehat{P})}-\fro - c_{\omega}$ with the hyperparameters $(\kappa, \lambda, \mu) = (.2, .4, .3, .1)$ achieving the lowest abstraction error.}
    \label{tab:downstream_multi}
\end{table}

\begin{table}
\begin{minipage}[b]{0.49\linewidth}
\resizebox{\columnwidth}{!}{%
\begin{tabular}{ccccc}
    \toprule
       Method & $\mathcal{D}$  & $\mathcal{C}$ & $e_{\jsd}(\tau)$ & $e_{\wass}(\tau)$  \\
    \midrule
    \cota{($\widehat{P}$)} & \fro  & $c_{\omega}$ & $0.010 \pm 0.005$ &$0.010 \pm 0.002$\\
     &  & $c_{\mathcal{H}}$ & $0.125 \pm 0.003$ &$0.011 \pm 0.003$\\
     \\
     & \jsd  & $c_{\omega}$ & $\bm{0.008 \pm 0.001}$&$\bm{0.009 \pm 0.001}$\\
     &   & $c_{\mathcal{H}}$ & $0.036 \pm 0.011$ &$0.016 \pm 0.002$\\
     \\
    \midrule
     \cota{($\widehat{\tau}$)}  & \fro  & $c_{\omega}$ & $0.013\pm 0.008$ &$0.171 \pm 0.001$\\
     &  & $c_{\mathcal{H}}$ & $0.096\pm 0.010$ &$0.175 \pm 0.001$\\
     \\
     & \jsd  & $c_{\omega}$ & $0.013\pm 0.007$ &$0.171 \pm 0.001$\\
     &   & $c_{\mathcal{H}}$ & $0.145\pm 0.008$ &$0.175 \pm 0.001$ \\
     \\
    \midrule
    \pairOT & --   & $c_{\omega}$    & $0.013 \pm 0.002$ &$0.011 \pm 0.002$\\
    & -- &    $c_{\mathcal{H}}$  & $0.093 \pm 0.004$ &$0.039 \pm 0.002$\\
    \midrule
    \mapOT{} & -- &  $c_{\omega}$    & $0.023 \pm 0.022$ &$0.147 \pm 0.001$\\
           & -- &  $c_{\mathcal{H}}$  & $0.169 \pm 0.022$ &$0.156 \pm 0.001$\\
    \midrule
    \baryOT{} & -- &  $c_{\omega}$    & $0.233 \pm 0.142$ &$0.067 \pm 0.042$\\
              & -- &  $c_{\mathcal{H}}$  & $0.323 \pm 0.074$ &$0.095 \pm 0.039$\\
    \bottomrule
        \end{tabular}}
        \captionsetup{skip=10pt}
\caption{{\texttt{Approximate} \cota{} complete abstraction error evaluation for the \textbf{STC}$_{\text{np}}$ example. The $\cota{(\widehat{P})}-\jsd - c_{\omega}$ setting returns the lower abstraction error for both metrics. The $\omega$-cost $c_{\omega}$ outperforms $c_{\mathcal{H}}$.}}
\label{tab:STC_NP_results_all}
\end{minipage}\hfill%
\begin{minipage}[b]{0.49\linewidth}
\resizebox{\columnwidth}{!}{%
\setlength{\tabcolsep}{10pt} 
\renewcommand{\arraystretch}{.97}
\begin{tabular}{ccccc}
    \toprule
    Method & $\mathcal{D}$  & $\mathcal{C}$ & $e_{\jsd}(\tau)$   & $e_{\wass}(\tau)$\\
    \midrule
     \cota{($\widehat{P}$)} & \fro  & $\omega$ & $0.264 \pm 0.001$&$0.064 \pm 0.001$\\
     &  & $\mathcal{H}$ & $0.241 \pm 0.003$ &$0.060 \pm 0.004$\\
     \\
     & \jsd  & $\omega$ & $0.259 \pm 0.006$&$\bm{0.051 \pm 0.002}$\\
     &   & $\mathcal{H}$ & $0.242 \pm 0.001$ &$\bm{0.051 \pm 0.001}$\\
     \\
    \midrule
     \cota{($\widehat{\tau}$)} & \fro  & $\omega$ & $0.248\pm 0.006$ &$0.135 \pm 0.001$\\
     &  & $\mathcal{H}$ & $\bm{0.227 \pm 0.005}$&$0.129 \pm 0.001$ \\
     \\
     & \jsd  & $\omega$ & $0.236 \pm 0.002$ &$0.130 \pm 0.005$\\
     &   & $\mathcal{H}$ & $\bm{0.229 \pm 0.006}$ &$0.129 \pm 0.001$\\
     \\
    \midrule
    \pairOT & --   & $\omega$    & $0.279 \pm 0.014$ &$0.091 \pm 0.005$\\
    & -- &    $\mathcal{H}$  & $0.242 \pm 0.002$ &$0.067 \pm 0.001$\\
    \midrule
    \mapOT{} & -- &  $\omega$    & $0.250 \pm 0.005$ &$0.140 \pm 0.001$\\
           & -- &  $\mathcal{H}$  & $0.229 \pm 0.004$ &$0.129 \pm 0.001$\\
    \midrule
    \baryOT{} & -- &  $\omega$    & $0.318 \pm 0.205$ &$0.104 \pm 0.061$\\
              & -- &  $\mathcal{H}$  & $0.272 \pm 0.212$ &$0.075 \pm 0.058$\\
    \bottomrule
        \end{tabular}}
        \captionsetup{skip=10pt}
\caption{\texttt{Approximate} \cota{} complete abstraction error evaluation for the \textbf{STC}$_{\text{p}}$ example. The $\cota{(\widehat{\tau})}-\fro - c_{\mathcal{H}}$ and $\cota{(\widehat{\tau})}-\jsd - c_{\mathcal{H}}$ settings returns the lower abstraction errors for $e_{\jsd}(\tau)$ and $\cota{(\widehat{P})}-\fro - c_{\omega}$ and $\cota{(\widehat{P})}-\jsd - c_{\mathcal{H}}$ settings returns the lower abstraction errors for $e_{\wass}(\tau)$. Table illustrates that $c_{\mathcal{H}}$ outperforms $c_{\omega}$ due to under-specification of the intervention set.}
\label{tab:STC_P_results_all}
\end{minipage}
\end{table}

\begin{table}
\begin{minipage}[b]{0.49\linewidth}
\resizebox{\columnwidth}{!}{%
\begin{tabular}{ccccc}
    \toprule
    Method & $\mathcal{D}$  & $\mathcal{C}$ & $e_{\jsd}(\tau)$   & $e_{\wass}(\tau)$\\
    \midrule
     \cota{($\widehat{P}$)} & \fro  & $c_{\omega}$ & $\bm{0.258 \pm 0.003}$&$\bm{0.044 \pm 0.001}$\\
     &  & $c_{\mathcal{H}}$ & $0.260 \pm 0.004$&$0.047 \pm 0.001$\\
     \\
     & \jsd  & $c_{\omega}$ & $0.285 \pm 0.014$&$0.045 \pm 0.001$\\
     &   & $c_{\mathcal{H}}$ & $0.285 \pm 0.014$ &$0.046 \pm 0.001$\\
     \\
    \midrule
     \cota{($\widehat{\tau}$)} & \fro  & $c_{\omega}$ & $0.259 \pm 0.006$ &$0.060 \pm 0.003$\\
     &  & $c_{\mathcal{H}}$ & $0.259 \pm 0.006$ &$0.060 \pm 0.003$\\
     \\
     & \jsd  & $c_{\omega}$ & $0.260 \pm 0.007$&$0.061 \pm 0.001$\\
     &   & $c_{\mathcal{H}}$ & $0.263 \pm 0.004$ &$0.061 \pm 0.001$\\
     \\
    \midrule
    \pairOT & --   & $c_{\omega}$    & $0.306 \pm 0.009$ &$0.045 \pm 0.001$\\
    & -- &    $c_{\mathcal{H}}$  & $0.387 \pm 0.002$ &$0.047 \pm 0.001$\\
    \midrule
    \mapOT{} & -- &  $c_{\omega}$    & $0.294 \pm 0.008$ &$0.054 \pm 0.001$\\
           & -- &  $c_{\mathcal{H}}$  & $0.350 \pm 0.005$ &$0.054 \pm 0.001$\\
    \midrule
    \baryOT{} & -- &  $c_{\omega}$    & $0.294 \pm 0.047$ &$0.044 \pm 0.003$\\
              & -- &  $c_{\mathcal{H}}$  & $0.414 \pm 0.040$ &$0.046 \pm 0.010$\\
    \bottomrule
        \end{tabular}}
        \captionsetup{skip=10pt}
\caption{{\texttt{Approximate} \cota{} complete abstraction error evaluation for the \textbf{LUCAS} example. The $\cota{(\widehat{P})}-\fro - c_{\omega}$ setting returns the lower abstraction error for both metrics. The $\omega$-cost $c_{\omega}$ outperforms $c_{\mathcal{H}}$.}}
\label{tab:LUCAS_results_all}
\end{minipage}\hfill%
\begin{minipage}[b]{0.49\linewidth}
\resizebox{\columnwidth}{!}{%
\setlength{\tabcolsep}{10pt} 
\renewcommand{\arraystretch}{.55}
\begin{tabular}{ccccc}
    \toprule
    Method & $\mathcal{D}$  & $\mathcal{C}$ & $e_{\jsd}(\tau)$   & $e_{\wass}(\tau)$\\
    \midrule
     \cota{($\widehat{P}$)} & \fro  & $c_{\omega}$ & $0.378$&$\bm{0.006}$\\
     &  & $c_{\mathcal{H}}$ & $0.263$ &$\bm{0.006}$\\
     \\
     & \jsd  & $c_{\omega}$ & $0.378$&$\bm{0.006}$\\
     &   & $c_{\mathcal{H}}$ & $0.263$ &$\bm{0.006} $\\
     \\
    \midrule
     \cota{($\widehat{\tau}$)} & \fro  & $c_{\omega}$ & $0.389$ &$0.053 $\\
     &  & $c_{\mathcal{H}}$ & $\bm{0.226}$&$0.053 $ \\
     \\
     & \jsd  & $c_{\omega}$ & $0.389$&$0.053 $\\
     &   & $c_{\mathcal{H}}$ & $\bm{0.226}$ &$0.053 $\\
     \\
    \midrule
    \pairOT & --   & $c_{\omega}$    & $0.430$ &$0.027 $\\
    & -- &    $c_{\mathcal{H}}$  & $0.263$ &$0.027 $\\
    \midrule
    \mapOT{} & -- &  $c_{\omega}$    & $0.408$ &$0.060 $\\
           & -- &  $c_{\mathcal{H}}$  & $0.228$ &$0.053 $\\
    \midrule     
    \baryOT{} & -- &  $c_{\omega}$    & $0.530 $ &$0.019 $\\
              & -- &  $c_{\mathcal{H}}$  & $0.335$ &$0.023 $\\
    \bottomrule
        \end{tabular}}
        \captionsetup{skip=10pt}
\caption{{\texttt{Approximate} \cota{} complete abstraction error evaluation for the \textbf{EBM} example. The $\cota{(\widehat{\tau})}-\fro - c_{\mathcal{H}}$ and $\cota{(\widehat{\tau})}-\jsd - c_{\mathcal{H}}$ settings returns the lower abstraction errors for $e_{\jsd}(\tau)$ and $\cota{(\widehat{P})}$ for all settings returns the lower abstraction errors for $e_{\wass}(\tau)$. Table illustrates that $c_{\mathcal{H}}$ outperforms $c_{\omega}$ due to under-specification of the intervention set collected from the labs.}}
\label{tab:EBM_results_all}
\end{minipage}
\end{table}

\begin{table}
    \begin{minipage}[b]{0.49\linewidth}
        \resizebox{\columnwidth}{!}{%
        \includegraphics{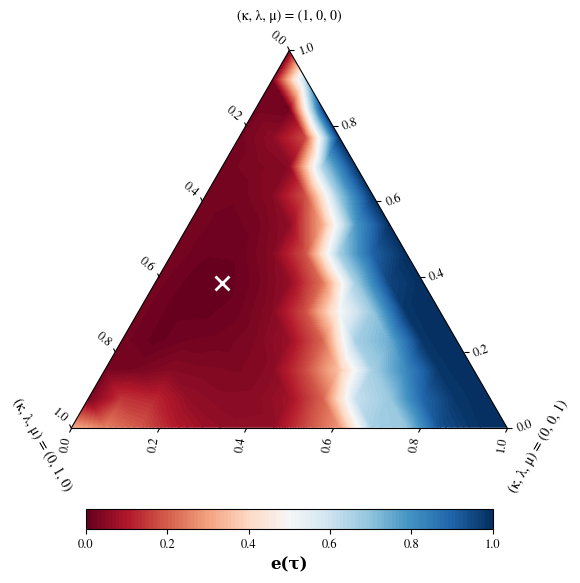}}
        \captionsetup{skip=10pt}
        \captionof{figure}{Effect of $\lambda$ for the \textbf{STC}$_{\text{np}}$ example in the \texttt{Approximate} \cota{} formulation. The ternary plot illustrates a grid-search amongst 100 convex combinations of $(\kappa, \lambda, \mu)$ for the $\cota{(\widehat{P})} - \fro - c_{\omega}$ setting. Values of $\lambda$ close to zero present high abstraction error, demonstrating the benefit of the \emph{do-calculus} constraints in the \ot{} problem. The minimum is reached at (.38, .46, .16) and is denoted with "$\bm{\times}$"}
        \label{fig:fro_ternary}
    \end{minipage}\hfill%
    \begin{minipage}[b]{0.49\linewidth}
        \resizebox{\columnwidth}{!}{%
        \includegraphics{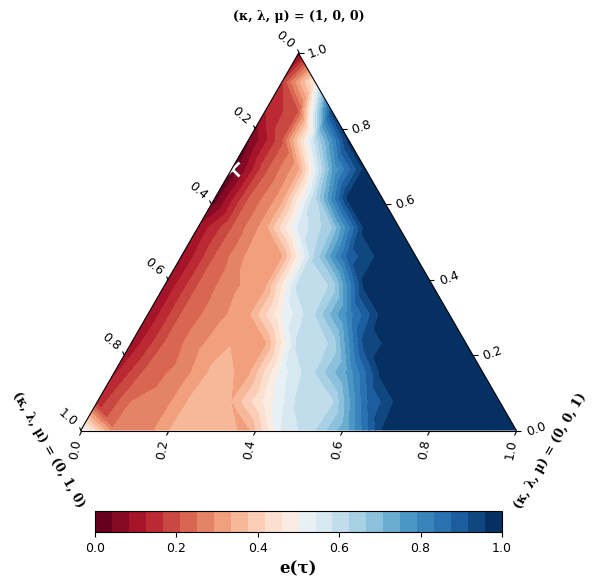}}
        \captionsetup{skip=10pt}
        \captionof{figure}{Effect of $\lambda$ for the \textbf{STC}$_{\text{np}}$ example in the \texttt{Approximate} \cota{} formulation. The ternary plot illustrates a grid-search amongst 100 convex combinations of $(\kappa, \lambda, \mu)$ for the $\cota{(\widehat{P})} - \jsd - c_{\omega}$ setting. Values of $\lambda$ close to zero present high abstraction error, demonstrating the benefit of the \emph{do-calculus} constraints in the \ot{} problem. The minimum is reached at (.61, .38, .01) and is denoted with "$\bm{\times}$"}
        \label{fig:jsd_ternary}
    \end{minipage}
\end{table}

\begin{table}
    \centering
    \begin{tabular}{cccc}
    \toprule
     Training set & Test set & \cite{zennaro2023jointly} & \cota{}\\
    \midrule \\
    LRCS[$CG\neq k$] & LRCS[$CG = k$] &  $1.86 \pm 1.75$ &  \bm{$1.40 \pm 1.39$} \\ 
    \midrule
    LRCS[$CG\neq k$] & LRCS[$CG = k$] & $0.22 \pm 0.26$ & \bm{$0.13 \pm 0.07$ }\\
    +WMG & & \\  
    \midrule
     LRCS[$CG\neq k$] & LRCS[$CG = k$] &  $1.22 \pm 0.95$ & \bm{ $0.85 \pm 0.81$ }\\ 
     +WMG[$CG\neq k$] & WMG[$CG = k$] &  \\
    \bottomrule
    \end{tabular} 
    \captionsetup{skip=10pt}
\caption{MSE of \texttt{Approximate} \cota{} and a SOTA \ca{} framework on a regression task for \textbf{EBM}. Augmenting
data via the learned abstraction reduces the average error in all different settings compared to the SOTA. We used $\cota{(\widehat{P})}-\fro - c_{\omega}$ with the hyperparameters $(\kappa, \lambda, \mu) = (.2, .5, .3)$ achieving the lowest abstraction error.}
    \label{tab:downstream_approx_cota}
\end{table}

\clearpage
\section{\dirag{s} and chains}\label{sec:dags_chains}
In this section we provide the complete \dirag{s} alongside their corresponding intervention posets and induced chains for all the scenarios that we considered and also visualise the operation of the $\omega$ map. For enhanced clarity and ease of navigation between different settings, we have included a concise table presented in \autoref{tab:vars_and_dags_roadmap}. To provide further insight, the table also offers an analytical description of the interpretation of endogenous variables within each example for both the base and abstracted models. 

\begin{figure}[H]
\centering
\begin{minipage}[b]{0.49\linewidth}
\centering
 \label{fig:stc_np_graphs}
\begin{tikzpicture}[shorten >=1pt, auto, node distance=1cm, thick, scale=0.8, every node/.style={scale=0.8}]
			\tikzstyle{node_style} = [circle,draw=black]
			\node[node_style] (S) at (0,0) {S};
			\node[node_style] (T) at (1.5,0) {T};
			\node[node_style] (C) at (3,0) {C};
			\draw[->]  (T) to (C);
			\draw[->]  (S) to (T);

                \node[node_style] (S_) at (.5,-2) {S'};
			\node[node_style] (C_) at (2.5,-2) {C'};
			\draw[->]  (S_) to (C_);

                \draw[->,dotted]  (1.5,-.7) to node[right]{$\tau$} (1.5,-1.5);
   
		\end{tikzpicture}
\end{minipage}%
\begin{minipage}[b]{0.49\linewidth}
\centering
 \label{fig:stc_np_graphs}
 \scalebox{1}{
\begin{tikzpicture}[shorten >=1pt, auto, node distance=1cm, thick, scale=0.7, every node/.style={scale=0.7}]
        \tikzstyle{node_style} = []
        
        \node[node_style,lowcolor] (null) at (0,0) {$\emptyset$};
        \node[node_style,lowcolor] (S0) at (-2,1) {$S=0$};
        \node[node_style,lowcolor] (S1) at (2,1) {$S=1$};
        \node[node_style,lowcolor] (S0T0) at (-2,2) {$S=0,T=1$};
        \node[node_style,lowcolor] (S1T0) at (.7,2) {$S=1,T=1$};
        \node[node_style,lowcolor] (S1T1) at (3.2,2) {$S=1,T=1$};

        \node[node_style,highcolor] (null_) at (7,.5) {$\emptyset$};
        \node[node_style,highcolor] (S0_) at (6,1.5) {$S'=0$};
        \node[node_style,highcolor] (S1_) at (8,1.5) {$S'=1$};
        
        \draw[->]  (null) to (S0);
        \draw[->]  (null) to (S1);
        \draw[->]  (S0) to (S0T0);
        \draw[->]  (S1) to (S1T0);
        \draw[->]  (S1) to (S1T1);

        \draw[->]  (null_) to (S0_);
        \draw[->]  (null_) to (S1_);
        
        \draw[->,dotted,red]  (null) to[bend right=10] (null_);
        \draw[->,dotted,red]  (S0) to (S0_);
        \draw[->,dotted,red]  (S1) to[bend right=10] (S1_);
        \draw[->,dotted,red]  (S0T0) to[bend left=30] (S0_);
        \draw[->,dotted,red]  (S1T0) to[bend left=20] (S1_);
        \draw[->,dotted,red]  (S1T1) to[bend left=10] (S1_);
        \end{tikzpicture}}
\end{minipage}
\captionsetup{skip=10pt}
\caption{\textit{Simple Lung Cancer} (\textbf{STC}) base (top) and abstracted (bottom) \dirag{s} alongside their equivalent posets $\intervsetbase$ and $\intervsetabst$ structure for the \textbf{STC}$_{\text{np}}$ variation. The red arrows represent the $\omega: \intervsetbase \to \intervsetabst$ map.}
\label{fig:stc_np_graphs}
\end{figure}

\begin{figure}[H]
\centering
\begin{minipage}[b]{0.49\linewidth}
\centering
 \label{fig:stc_p_graphs}
\begin{tikzpicture}[shorten >=1pt, auto, node distance=1cm, thick, scale=0.8, every node/.style={scale=0.8}]
			\tikzstyle{node_style} = [circle,draw=black]
			\node[node_style] (S) at (0,0) {S};
			\node[node_style] (T) at (1.5,0) {T};
			\node[node_style] (C) at (3,0) {C};
			\draw[->]  (T) to (C);
			\draw[->]  (S) to (T);

                \node[node_style] (S_) at (.5,-2) {S'};
			\node[node_style] (C_) at (2.5,-2) {C'};
			\draw[->]  (S_) to (C_);

                \draw[->,dotted]  (1.5,-.7) to node[right]{$\tau$} (1.5,-1.5);
   
		\end{tikzpicture}
\end{minipage}%
\begin{minipage}[b]{0.49\linewidth}
\centering
\label{fig:stc_p_graphs} 
\scalebox{1}{
\begin{tikzpicture}[shorten >=1pt, auto, node distance=1cm, thick, scale=0.7, every node/.style={scale=0.7}]
        \tikzstyle{node_style} = []
        
        \node[node_style,lowcolor] (null) at (0,0) {$\emptyset$};
        \node[node_style,lowcolor] (T0) at (-2,1) {$T=0$};
        \node[node_style,lowcolor] (T1) at (2,1) {$T=1$};

        \node[node_style,highcolor] (null_) at (7,.5) {$\emptyset$};
        \node[node_style,highcolor] (C0_) at (6,1.5) {$C'=0$};
        \node[node_style,highcolor] (C1_) at (8,1.5) {$C'=1$};
        
        \draw[->]  (null) to (T0);
        \draw[->]  (null) to (T1);

        \draw[->]  (null_) to (C0_);
        \draw[->]  (null_) to (C1_);
        
        \draw[->,dotted,red]  (null) to[bend right=10] (null_);
        \draw[->,dotted,red]  (T0) to[bend left=10] (C0_);
        \draw[->,dotted,red]  (T1) to[bend right=10] (C1_);
        \end{tikzpicture}}
\end{minipage}
\captionsetup{skip=10pt}
\caption{\textit{Simple Lung Cancer} (\textbf{STC}) base (top) and abstracted (bottom) \dirag{s} alongside their equivalent posets $\intervsetbase$ and $\intervsetabst$ structure in the \textbf{STC}$_{\text{p}}$ variation. The red arrows represent the $\omega: \intervsetbase \to \intervsetabst$ map.}
\label{fig:stc_p_graphs}
\end{figure}

\begin{figure}[H]
\centering
\begin{minipage}[b]{0.49\linewidth}
\centering
 \label{fig:lucas_graphs}
\begin{tikzpicture}[shorten >=1pt, auto, node distance=1cm, thick, scale=0.8, every node/.style={scale=0.8}]
			\tikzstyle{node_style} = [circle,draw=black]
			\node[node_style] (AN) at (-5,0) {AN};
			\node[node_style] (SM) at (-3,0) {SM};
			\node[node_style] (LC) at (-1,0) {LC};
                \node[node_style] (CO) at (1,0) {CO};
			\node[node_style] (AL) at (3,0) {AL};
                \node[node_style] (PP) at (-3,2) {PP};
			\node[node_style] (GE) at (-1,2) {GE};
                \node[node_style] (FA) at (1,2) {FA};
			\draw[->]  (AN) to (SM);
			\draw[->]  (PP) to (SM);
                \draw[->]  (SM) to (LC);
			\draw[->]  (GE) to (LC);
                \draw[->]  (LC) to (FA);
			\draw[->]  (LC) to (CO);
                \draw[->]  (AL) to (CO);

                \node[node_style] (LC_) at (-1,-2) {LC'};
			\node[node_style] (EN_) at (-4,-2) {EN'};
                \node[node_style] (GE_) at (2,-2) {GE'};
			\draw[->]  (GE_) to (LC_);
                \draw[->]  (EN_) to (LC_); 

                \draw[->,dotted]  (-1,-0.5) to node[right]{$\tau$} (-1,-1.5);
   
		\end{tikzpicture}
\end{minipage}%
\begin{minipage}[b]{0.49\linewidth}
\centering
\label{fig:lucas_graphs}
\scalebox{1}{
\begin{tikzpicture}[shorten >=1pt, auto, node distance=1cm, thick, scale=0.7, every node/.style={scale=0.7}]
        \tikzstyle{node_style} = []
        
        \node[node_style,lowcolor] (null) at (-3,-3) {$\emptyset$};
        \node[node_style,lowcolor] (AN0) at (-5,-1) {$AN=0$};
        \node[node_style,lowcolor] (GE1) at (-3,-1) {$GE=1$};
        \node[node_style,lowcolor] (AL0) at (-1,-1) {$AL=0$};
        \node[node_style,lowcolor] (AN0PP0) at (-5,0) {$AN,PP=0,0$};
        \node[node_style,lowcolor] (AN0PP0SM0) at (-7,1) {$AN,PP,SM=0,0,0$};
        \node[node_style,lowcolor] (AN0PP0SM1) at (-2.5,1) {$AN,PP,SM=0,0,1$};

        \node[node_style,highcolor] (null_) at (2,-3) {$\emptyset$};
        \node[node_style,highcolor] (EN_0) at (0.3,0.3) {$EN'=0$};
        \node[node_style,highcolor] (EN_1) at (1.2,1.2) {$EN'=1$};
        \node[node_style,highcolor] (GE_0) at (2.2,0.3) {$GE'=0$};
        \node[node_style,highcolor] (GE_1) at (3.6,1.2) {$GE'=1$};
        
        \draw[->]  (null) to (AN0);
        \draw[->]  (null) to (GE1);
        \draw[->]  (null) to (AL0);
        \draw[->]  (AN0) to (AN0PP0);
        \draw[->]  (AN0PP0) to (AN0PP0SM0);
        \draw[->]  (AN0PP0) to (AN0PP0SM1);

        \draw[->]  (null_) to (EN_0);
        \draw[->]  (null_) to (EN_1);
        \draw[->]  (null_) to (GE_0);
        \draw[->]  (null_) to (GE_1);
        
        \draw[->,dotted,red]  (null) to[bend left=10] (null_);
        \draw[->,dotted,red]  (AN0) to[bend left=10] (EN_0);
        \draw[->,dotted,red]  (AN0PP0) to[bend left=20] (EN_0);
        \draw[->,dotted,red]  (AN0PP0SM0) to[bend left=20] (EN_0);
        \draw[->,dotted,red]  (AN0PP0SM1) to[bend left=30] (EN_0);
        \draw[->,dotted,red]  (GE1) to[bend right=10] (GE_1);
        \draw[->,dotted,red]  (AL0) to[bend right=10] (GE_0);
        
        \end{tikzpicture}}
\end{minipage}
\captionsetup{skip=10pt}
\caption{(\textbf{LUCAS}) base (top) and abstracted (bottom) \dirag{s} alongside their equivalent posets $\intervsetbase$ and $\intervsetabst$ structure. The red arrows represent the $\omega: \intervsetbase \to \intervsetabst$ map.}
\label{fig:lucas_graphs}
\end{figure}

\begin{figure}[H]
\centering
\begin{minipage}[b]{0.49\linewidth}
\centering
 \label{fig:ebm_graphs}
\begin{tikzpicture}[shorten >=1pt, auto, node distance=1cm, thick, scale=0.7, every node/.style={scale=0.7}]
			\tikzstyle{node_style} = [circle,draw=black]
			\node[node_style] (CG) at (1.5,0) {CG};
			\node[node_style] (ML1) at (3,1) {ML1};
                \node[node_style] (ML2) at (3,-1) {ML2};
                \draw[->]  (CG) to (ML1);
			\draw[->]  (CG) to (ML2);

                \node[node_style] (CG_) at (1.5,-3) {CG'};
                \node[node_style] (ML_) at (3,-3) {ML'};
			\draw[->]  (CG_) to (ML_);

                \draw[->,dotted]  (2.2,-1.2) to node[right]{$\tau$} (2.2,-2.2);
   
		\end{tikzpicture}
\end{minipage}%
\begin{minipage}[b]{0.49\linewidth}
\centering
 \label{fig:ebm_graphs}
 \scalebox{1}{
\begin{tikzpicture}[shorten >=1pt, auto, node distance=1cm, thick, scale=0.7, every node/.style={scale=0.7}]
        \tikzstyle{node_style} = []
        
        \node[node_style,lowcolor] (null) at (-2,0) {$\emptyset$};
        \node[node_style,lowcolor] (CG75) at (-5,1) {$CG=75$};
        \node[node_style,lowcolor] (CG110) at (-3,1) {$CG=110$};
        \node[node_style,lowcolor] (CG180) at (-1,1) {$CG=180$};
        \node[node_style,lowcolor] (CG200) at (1,1) {$CG=200$};

        \node[node_style,highcolor] (null_) at (6,0) {$\emptyset$};
        \node[node_style,highcolor] (CG75_) at (3,1) {$CG'=75$};
        \node[node_style,highcolor] (CG100_) at (5,1) {$CG'=100$};
        \node[node_style,highcolor] (CG200_) at (7,1) {$CG'=200$};
        
        \draw[->]  (null) to (CG75);
        \draw[->]  (null) to (CG110);
        \draw[->]  (null) to (CG180);
        \draw[->]  (null) to (CG200);

        \draw[->]  (null_) to (CG75_);
        \draw[->]  (null_) to (CG100_);
        \draw[->]  (null_) to (CG200_);
        
        \draw[->,dotted,red]  (null) to[bend right=10] (null_);
        \draw[->,dotted,red]  (CG75) to[bend left=10] (CG75_);
        \draw[->,dotted,red]  (CG110) to[bend left=20] (CG100_);
        \draw[->,dotted,red]  (CG180) to[bend right=20] (CG200_);
        \draw[->,dotted,red]  (CG200) to[bend right=10] (CG200_);
        
        \end{tikzpicture}}
\end{minipage}
\captionsetup{skip=10pt}
\caption{\textit{Electric Battery Manufacturing} (\textbf{EBM}) base (top) and abstracted (bottom) \dirag{s} alongside their equivalent posets $\intervsetbase$ and $\intervsetabst$ structure. The red arrows represent the $\omega: \intervsetbase \to \intervsetabst$ map.}
\label{fig:ebm_graphs}
\end{figure}

\vspace{5mm}
\begin{table}[H]
    \centering
    \setlength{\tabcolsep}{8pt} 
\renewcommand{\arraystretch}{1.5}
    \begin{tabular}{|c|c|c|c|}
    \hline
    \textbf{\Large{Example}} & \textbf{\Large{Figure}} &\Large $\mathbf{X}$  & \Large $\mathbf{X^{\prime}}$\\
   \hline
   \hline
     \textbf{STC} & \autoref{fig:stc_np_graphs} & \textbf{S}: Smoking, &\textbf{S$'$}: Smoking, \\
                  & \autoref{fig:stc_p_graphs} & \textbf{T}: Tar,     &\textbf{C$'$}: Cancer  \\
                  &        & \textbf{C}:Cancer   &                       \\
    \hline
     \textbf{LUCAS} &  \autoref{fig:lucas_graphs} & \textbf{AN}: Anxiety, \textbf{SM}: Smoking,        & \textbf{EN$'$}: Environment,  \\
                    &        & \textbf{GE}: Genetics, \textbf{PP}: Peer Pressure, &  \textbf{LC$'$}: Lung Cancer, \\
                    &        & \textbf{LC}: Lung Cancer, \textbf{FA}: Fatigue,    & \textbf{GE$'$}: Genetics     \\
                    &        & \textbf{CO}: Coughing, \textbf{AL}: Allergy        &                              \\
     \hline
      \textbf{EBM} & \autoref{fig:ebm_graphs} & \textbf{CG}: Comma Gap (lab 1),               & \textbf{CG$'$}: Comma Gap (lab 2),    \\
                   &        & \textbf{ML1}: Mass loading position 1 (lab 1), & \textbf{ML$'$}: Mass loading (lab 2) \\
                   &        & \textbf{ML1}: Mass loading position 2 (lab 1)  &                                      \\
    \hline
    \end{tabular}
    \captionsetup{skip=10pt}
    \caption{Analytical interpretation of the base $\scmbase$ and abstracted $\scmabst$ model's endogenous variables.}
    \label{tab:vars_and_dags_roadmap}
\end{table}

\clearpage
\section{Optimal Transport}\label{sec:ot_appendix_content}
Optimal Transport (\ot{}) theory as surveyed in \cite{villani2009optimal,santambrogio2015optimal} provides a mathematical framework for systematically mapping one probability measure $\mu$ to another $\nu$ by looking amongst the set of all possible ways to transport the mass from one \emph{source} distribution to a \emph{target} one and selecting the one which minimizes a cost function. The seminal work of \cite{peyre2019computational} surveys computational algorithms to solve \ot{} problems in practice. Overall, \ot{} offers a versatile and powerful tool for tackling complex and diverse problems across various fields, from image processing to economics.

\subsection{General Measures}

\paragraph{Monge formulation} The initial problem formulation was given by Gaspard \cite{monge1781memoire} and states the following: For two arbitrary measures $\mu$, $\nu$ on the Radon spaces\footnote{Radon space is a separable metric space such that any probability measure on it is a Radon measure} $\mathcal{X}$ and $\mathcal{Y}$ respectively, the \emph{Monge problem} seeks to find a \textit{map} $T^*:\mathcal{X}\mapsto \mathcal{Y}$ such that:
\begin{equation}\label{eq:monge_general}
T^* = \inf_{T}\Big\{ \int_{\mathcal{X}}c(x,T(x))d\mu(x):T_{\#}\mu = \nu\Big\}
\end{equation}
where $T_{\#}\mu$ is the pushforward function\footnote{Let $(X_1, \Sigma_1, \mu)$ be a measure space, $(X_2, \Sigma_2)$ a measurable
space, and $f : X_1 \to X_2$ a measurable map. Then the following function $\nu$
on $S_2$ is the \emph{pushforward} measure: $\nu(B) = \mu(f^{-1}(B))$ for $B\in \Sigma_2$. We write $f_{\#}\mu = \nu$.} and $c:\mathcal{X} \times \mathcal{Y}\mapsto \mathbb{R}$ is the cost function representing the cost of moving a unit mass from a location $x$ to a location $y$.
If such a $T^*$ exists and attains the infimum then this is called the \emph{optimal transport map}.

\begin{figure}[h]%
    \centering
    \includegraphics[width=9cm]{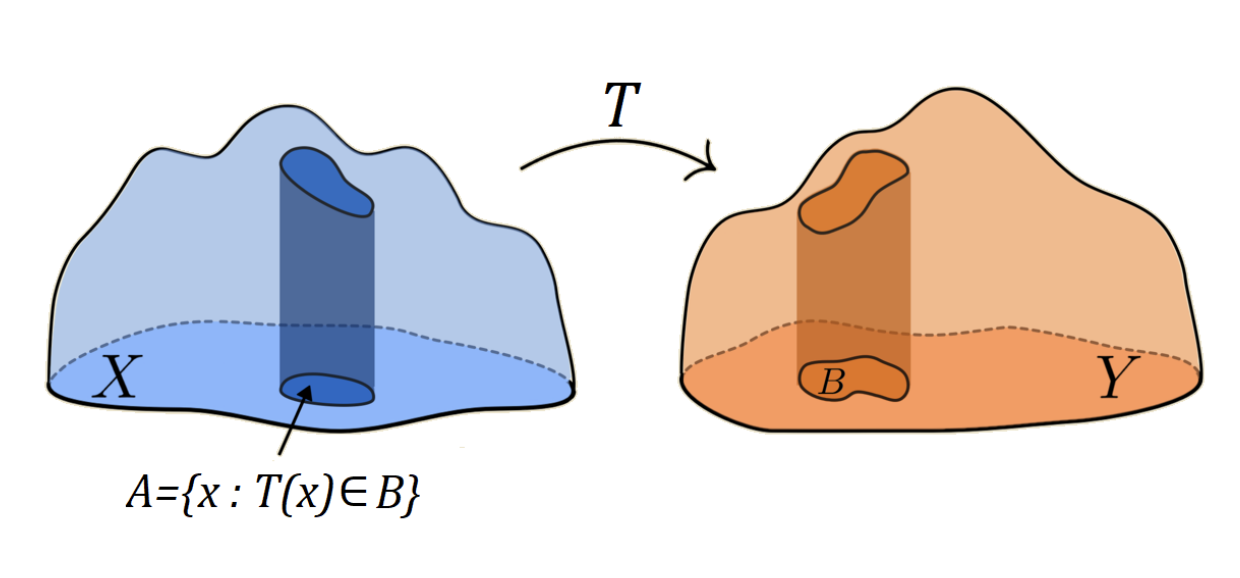} 
    \caption{Monge transport map (source: \citep{kolourisignalprocessing} Figure 1)}%
    \label{fig:example}%
\end{figure}

\paragraph{Kantorovic formulation} In the initial problem formulation by Monge the map $T^*$ may not always exist, for example, when $\mu$ is a Dirac measure but $\nu$ is not there is no map to attain the infimum of the optimization problem above.

For this reason, \cite{kantorovich1942} formulation of the problem relaxes the deterministic approach of Monge and introduces the probabilistic transport idea, which allows the execution of mass splitting from a source toward several targets. Specifically, the Kantorovic problem seeks to find a joint probability measure over the space $\mathcal{X} \times \mathcal{Y}$ or \textit{probabilistic coupling}, which solves the following optimization problem:
\begin{equation}\label{eq:kantorovic_general}
P^\star=\inf_{P}\left\{\int_{\mathcal{X}\times \mathcal{Y}}c(x,y)dP(x,y) | P \in \mathcal{U}(\mu, \nu)\right\}
\end{equation}
where $\mathcal{U}(\mu, \nu)$ is the collection of all probability measures on $\mathcal{X}\times \mathcal{Y}$ with marginals $\mu$ on $\mathcal{X}$ and $\nu$ on $\mathcal{Y}$. Namely, 
\begin{equation}
    \mathcal{U}(\mu, \nu) = \big\{P \in\mathcal{M}_+^1(\mathcal{X}\times \mathcal{Y}): \rho_{\mathcal{X\#}}P = \mu, ~~  \rho_{\mathcal{Y\#}}P = \nu \big\}
\end{equation}
where, $\rho_{\mathcal{X\#}}$ and $\rho_{\mathcal{Y\#}}$ are the pushforwards of the projections $\rho_{\mathcal{X}}(x,y)=x$ and $\rho_{\mathcal{Y}}(x,y)=y$.

This is an infinite-dimensional linear program over a space of measures. It is clear that this relaxed version of the problem is easier to work with since instead of looking for a map which that associates to each point $x_i \in \mathcal{X}$ a \textbf{single} point $y_i = T(x_i) \in \mathcal{Y}$, we are looking for a probability measure with the only constraint to preserve the marginals.

The Kantorovic formulation optimal transport computes the \emph{p-Wasserstein} distance metric $Wp(\cdot, \cdot)$ between two probability distributions $\mu$ and $\nu$:
\begin{equation}
W_p^p(\mu,\nu) = \inf_{P \in \mathcal{U}(\mu, \nu)}\int_{\mathcal{X} \times \mathcal{Y}}||x-y||^pdP(x,y), ~~ p \geq 1
\end{equation}
\paragraph{Remarks}
\begin{itemize}
    \item Let $T$ be a transport map between $\mu$ and $\nu$, and define $P_T=(id,T)_{\#}\mu$. Then, $P_T \in \mathcal{U}(\mu, \nu)$ is a transport plan between $\mu$ and $\nu$.
    \item Let $\mathcal{X}, \mathcal{Y}$ be two compact spaces, and $c : \mathcal{X} \times \mathcal{Y} \mapsto \mathbb{R}\cup\{+\infty\}$ be a lower semi-continuous cost function, which is bounded from below. Then Kantorovich’s problem admits a minimizer \citep{Thorpe2017IntroductionTO}.
\end{itemize}

\paragraph{Monge–Kantorovich equivalence for general measures}
The following theorem ensures that under some relatively simple conditions the Monge problem is feasible, meaning that the infimum of \cref{eq:monge_general} can be attained and thus, the Kantorovich and Monge formulations are equivalent.
\begin{theorem*}[\citep{brenierthm}]\label{def:brenier}
For Radon spaces $\mathcal{X}$ and $\mathcal{Y}$ with arbitrary measures $\mu, \nu$ respectively, if at least one of the two input measures, say $\mu$ has a density $\rho_{\mu}$ with respect to the Lebesgue measure, then there exists a unique  (up to an additive constant) convex function $\phi:\mathbb{R}^d \mapsto \mathbb{R}$ such that $\nabla \phi$ pushes forward $\mu$ onto $\nu$. In other words, there exists a deterministic coupling $P^\star$ as follows:
\begin{equation}
    dP^\star(x,y)=d\mu(x)\delta_{\nabla \phi(x)}(y)
\end{equation}
Furthermore, if $\mathcal{X}=\mathcal{Y}=\mathbb{R}^d$ and $c(x,y)=||x-y||^2$ then the optimal $P^\star$ in the Kantorovic formulation is unique and is supported on the graph $(x,T(x))$ of a Monge map $T:\mathbb{R}^d \to \mathbb{R}^d$. More formally,
\begin{equation}
    P = (id, T)_{\#\mu} \iff \forall h \in \mathcal{C}(\mathcal{X}\times \mathcal{Y}), ~~ \int_{\mathcal{X}\times \mathcal{Y}}h(x,y)dP(x,y)=\int_{\mathcal{X}}h(x,T(x))d\mu(x)
\end{equation}
This means that the map $T$ is uniquely defined as the gradient of a unique convex function $\phi$ such that $T(x)=\nabla \phi(x)$, where $(\nabla \phi)_{\#}\mu=\nu$.
\end{theorem*}
The two main conclusions from Brenier's theorem are the following:
\begin{itemize}
    \item In the setting of $\mathcal{W}^2$ with no-singular densities, the Monge problem \cref{eq:monge_general} and its Kantorovich relaxation \cref{eq:kantorovic_general} are equivalent (the relaxation is tight).
    \item An optimal transport map (Monge map) must be the gradient of a convex function. Namely, if $\phi:\mathbb{R}^d \to \mathbb{R}$ convex and $(\nabla \phi)_{\#}\mu=\nu$, then $T(x)=\nabla \phi(x)$ and \begin{equation}
        T^\star=\int_{\mathcal{X}}||x-\nabla\phi(x)||^2d\mu(x)
    \end{equation}
\end{itemize}
Various works extended the existence and uniqueness of Monge maps including strictly convex and super-linear costs.

\paragraph{Probabilistic interpretation} Both Monge and Kantorovich formulations can be reinterpreted through the prism of random variables \citep{mapestimation2018flamary}, \cite{peyre2019computational}. Consider two complete metric spaces $\mathcal{X}$ and $\mathcal{Y}$ and random variables $X$ and $Y$. We denote $X \sim \mu$ to say that $X$ is distributed according to the probability measure $\mu$. We can now restate both formulations of the optimal transport problem. Specifically, consider a cost function $c: \mathcal{X} \times \mathcal{Y} \to \mathbb{R}_{\geq 0}$ and two random variables $X \sim \mu$ and $Y \sim \nu$ taking values in $\mathcal{X}$ and $\mathcal{Y}$ respectively.

\emph{Monge formulation:} Find a map $T:\mathcal{X} \to \mathcal{Y}$ which transports the mass from $\mu$ to $\nu$ while minimizing the transportation cost,
\begin{align}
    \inf_{T} \mathbb{E}_{X \sim \mu}\left[c(X,T_{\#\mu}(X))\right] ~~ \text{s.t} ~~ T_{\#\mu}(X) \sim Y
\end{align}
\emph{Kantorovic formulation:} 
Find a coupling $(X,Y) \sim P$ which minimizes the transportation cost and asserts that $P$ has marginals equals to $\mu$ and $\nu$,
\begin{align}
    \inf_{P} \mathbb{E}_{(X,Y) \sim P}\left[c(X,Y)\right] ~~ \text{s.t} ~~ X \sim \mu,~~ Y \sim \nu
\end{align}

\subsection{Discrete Measures}
In this section, we introduce the notations and the formulation of \ot{} between  discrete distributions. A probability vector is any element $\bm{\alpha}$ that belongs to the probability simplex $\Sigma_k$:
\begin{align}
    \Sigma_k := \left\{\bm{\alpha} \in \mathbb{R}_{\geq 0}^k : \sum_{i=1}^k\alpha_i=1\right\}
\end{align}
A \emph{discrete measure} $\mu$ with weights $\bm{\alpha}$ and points $x_1,...,x_k \in \mathcal{X} \subset \mathbb{R}^d$ is defined as:
\begin{equation}
    \mu = \sum_{i=1}^k\alpha_i\delta_{x_i}
\end{equation}
where $\delta_{x}$ is the delta Dirac at position $x$. This measure is a probability measure if $\mu \in \Sigma_k$.

We are now going to restate the Monge-Kantorovic formulations in the cases of discrete measures. Consider $\mathcal{X}= \{x_i\}_{i=1}^M\subset \mathbb{R}^d$ and $\mathcal{Y}= \{y_j\}_{j=1}^N\subset \mathbb{R}^d$ with respective (probability) weights $\bm{\alpha} \in \Sigma_M, \bm{\beta} \in \Sigma_N$. 
Thus, we have the discrete probability measures:
\begin{equation}
    \mu = \sum_{i=1}^M\alpha_i\delta_{x_i} ~~ \text{and} ~~ \nu = \sum_{j=1}^N\beta_j\delta_{y_j}
\end{equation}
Finally, assuming that the cost of transporting a unit of mass from $x_i$ to $y_j$ is $c(x_i,y_j)$ where $c:\mathcal{X} \times \mathcal{Y} \to \mathbb{R}_{\geq 0}$ is the \emph{cost function}, this induces a \emph{cost matrix} $C_{ij}=c(x_i,y_j)$.

\paragraph{Monge formulation for discrete measures} 
The Monge formulation of \ot{} then aims to find a map $T^{\star}:\mathcal{X} \to \mathcal{Y}$ that push-forwards $\mu$ onto $\nu$, by assigning to each $x_i$ a single point $y_j$. Formally,
\begin{equation}\label{eq:monge_app}
T^{\star} = \ot^M_c(\mu,\nu)=\argmin_{T:~ T_{\#}\mu = \nu}\sum_{i=1}^Mc(x_i,T(x_i))
\end{equation}

\paragraph{Kantorovic formulation for discrete measures} Following the probabilistic transport approach of \cref{eq:kantorovic_general} for the general measures, the Kantorovich problem for discrete measures solves the following optimization problem in the form of a convex linear program: 
\begin{align}\label{eq:kantorovic_app}
P^{\star} = \ot^K_c(\mu,\nu)=\argmin_{P\in\mathcal{U(\mu,\nu)}}\left<C,P\right>=\argmin_{P\in\mathcal{U(\mu,\nu)}}\sum_{i=1, j=1}^{M, N} C_{ij} P_{ij}
\end{align}
where $\left<\cdot,\cdot\right>$ denotes the Frobenius inner product, expressing the total transportation cost, $C \in  \mathbb{R}_{\geq 0}^{M \times N}$ is the cost matrix and $\mathcal{U}(\mu,\nu)$ is the set of joint probability measures with marginals $\mu$ and $\nu$ and called the \emph{transport polytope} or \emph{coupling set}. In particular, the transport polytope is a convex polytope defined as follows:
\begin{align}
\mathcal{U}(\mu,\nu)= \left\{P\in \mathbb{R}_{\geq 0}^{M \times N}: P\onevector{N}=\mu, P^\transpose \onevector{M}=\nu \right\}  = \left\{P\in \mathbb{R}^{M \times N}:\sum_{j=1}^NP_{ij}=\mu, ~\sum_{i=1}^MP_{ij}=\nu \right\}
\end{align}

In \cref{fig:example} we provide a schematic viewed of input measures $(\mu, \nu)$ and a coupling $\mathcal{U}(\mu,\nu)$ encountered in the case of discrete measures for the Kantorovich \ot{} formulation for the square euclidean cost $c(x,y)=||x-y||^2$.
\begin{figure}[t]%
    \centering
    \includegraphics[width=14cm]{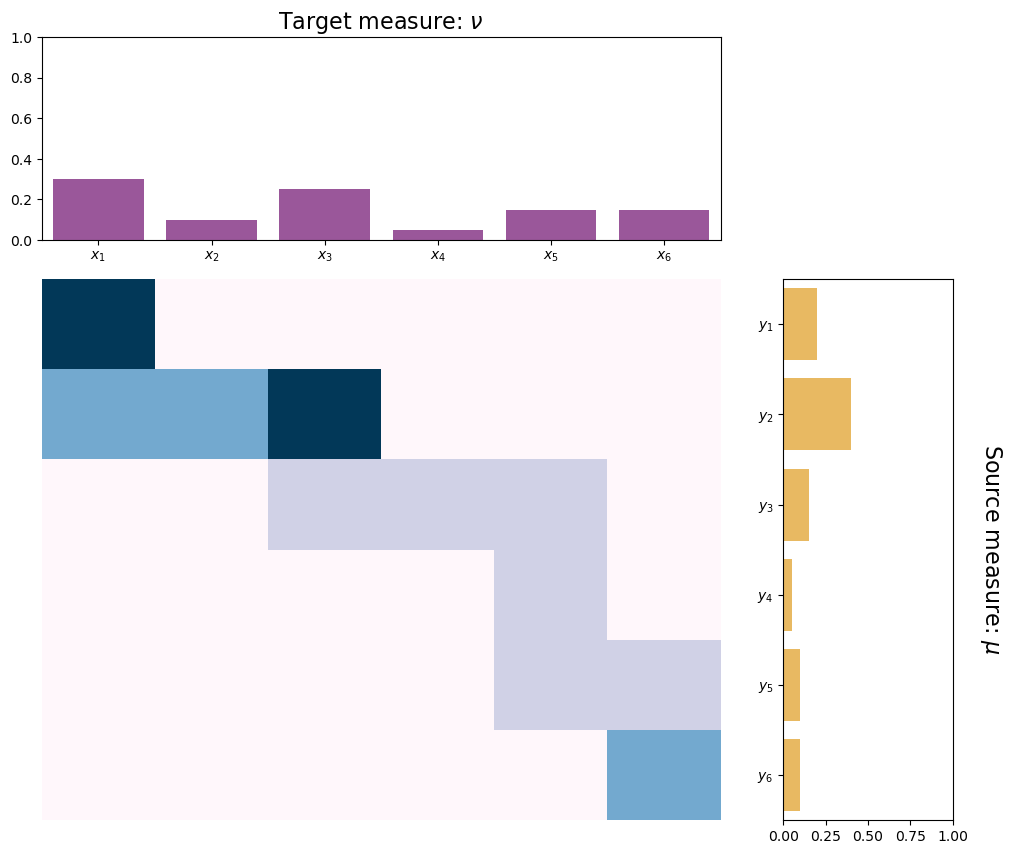}
    \caption{Kantorovic $\left(\ot_c^{K}(\mu,\nu)\right)$ optimal coupling for two input measures $\mu,\nu$. The optimal coupling $P^\star$ belongs to the transport polytope $\mathcal{U}(\mu,\nu)$ and thus preserves the marginals and the total mass, i.e. $\sum_{j}P^\star_{ij}=\mu, ~\sum_{i}P^\star_{ij}=\nu$ and $\sum_{i,j}P^\star_{ij}=1$.}%
    \label{fig:example}%
\end{figure}

\begin{figure}[t]%
    \centering
    \includegraphics[width=14cm]{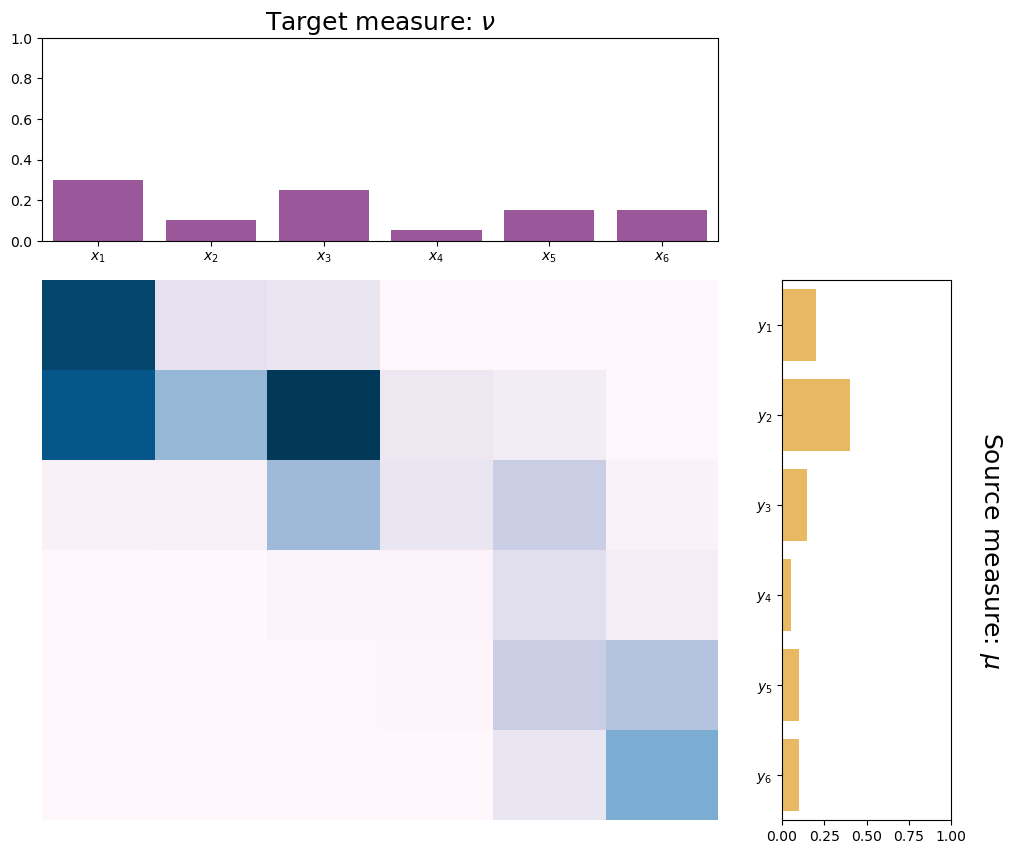}
    \caption{Kantorovic Entropic \ot{} $\left(\ot_c^{K}(\mu,\nu)_{\epsilon}\right)$ optimal coupling for two input measures $\mu,\nu$ and $\epsilon>0$. The optimal coupling $P^\star$ belongs to the transport polytope $\mathcal{U}(\mu,\nu)$ and thus preserves the marginals and the total mass, i.e. $\sum_{j}P^\star_{ij}=\mu, ~\sum_{i}P^\star_{ij}=\nu$ and $\sum_{i,j}P^\star_{ij}=1$.}%
    \label{fig:example1}%
\end{figure}

\paragraph{Entropic Optimal Transport}
Traditional \ot{}, while being a powerful tool, often encounters computational and statistical challenges in high-dimensional spaces. The introduction of entropy into this framework \citep{peyre2019computational} offers an organic solution that facilitates scalability and computational tractability through specific algorithms like Sinkhorn \citep{cuturi2013sinkhorn}. Overall, Entropic \ot{} leverages the principles of information theory, allowing for a more flexible and robust approach to the transportation problem between distributions. Specifically, for discrete measures, Entropic \ot{} solves the following optimisation problem:
\begin{align}\label{eq:entropic_ot_app}
P^{\star} = \ot_c^{K}(\mu,\nu)_{\epsilon}=\argmin_{P\in\mathcal{U(\mu,\nu)}}\left<C,P\right>-\epsilon\mathcal{H}(P) =\argmin_{P\in\mathcal{U(\mu,\nu)}}\sum_{i=1, j=1}^{M, N} C_{i, j} P_{i, j} -\epsilon\mathcal{H}(P) 
\end{align}
where $\epsilon>0$ a trade-off parameter and $\mathcal{H}(P)$ is the discrete entropy of a coupling matrix $P$ is defined as:
\begin{align}
    \mathcal{H}(P) := -\sum_{ij}P_{ij}(\log(P_{ij}-1))
\end{align}
The idea behind Entropic regularization in optimal transport involves employing a regularization function to derive approximate solutions to the original transport problem of \cref{eq:kantorovic_app}.
\paragraph{Remarks \citep{peyre2019computational}}
\begin{itemize}
    \item $ \ot_c^{K}(\mu,\nu)_{\epsilon}\xrightarrow{\epsilon\to 0}\ot_c^{K}(\mu,\nu)$
    \item $\ot_c^{K}(\mu,\nu)_{\epsilon} \xrightarrow{\epsilon\to +\infty} \mu\otimes\nu = \mu \nu^\top$
\end{itemize}
Finally, it is worth mentioning that, given the strong concavity of $\mathcal{H}$, the objective in \cref{eq:entropic_ot_app} becomes an $\epsilon$-strongly convex function, ensuring the optimization problem $\ot_c^{K}(\mu,\nu)_{\epsilon}$ has a unique solution. In \cref{fig:example1} we provide again the optimal coupling $\mathcal{U}(\mu,\nu)$ encountered in the case of discrete measures for the Entropic \ot{} formulation for the square euclidean cost $c(x,y)=||x-y||^2$.
\end{appendix}
\end{document}